\documentclass{article} 
\usepackage{iclr2027_conference,times}

\usepackage{amsmath,amsfonts,bm}

\def\eqref#1{equation~\ref{#1}}

\def\1{\bm{1}}

\DeclareMathAlphabet{\mathsfit}{\encodingdefault}{\sfdefault}{m}{sl}
\SetMathAlphabet{\mathsfit}{bold}{\encodingdefault}{\sfdefault}{bx}{n}

\usepackage{hyperref}
\usepackage{url}
\usepackage{comment}
\usepackage{graphicx}
\usepackage{booktabs}
\usepackage{multirow}
\usepackage{makecell}
\usepackage[table]{xcolor}
\usepackage{amsmath} 
\usepackage{wrapfig}
\usepackage{adjustbox}
\usepackage{caption}
\usepackage{algorithm}
\usepackage{algpseudocode}
\usepackage{float}\usepackage{xurl}
\usepackage{tikz}
\usepackage{pgfplots}
\usepackage{capt-of}
\usepackage[most]{tcolorbox}
\usepackage{xcolor}
\usepackage{amssymb}
\usepackage{colortbl}
\usepackage{tabularx}

\title{\framework{}: Long-Horizon Simulation of Everyday Human Behavior with Evolving States}

\author{%
\normalfont
\textbf{Ziwei Gong*}$^{1}$ \quad
\textbf{Xiangyang Yin*}$^{2}$ \quad
\textbf{Thai Quoc On}$^{1}$ \quad 
\textbf{Lin Ai}$^{3}$ \quad 
\textbf{Julia Hirschberg}$^{1}$ \quad 
\textbf{Haotong Yang}$^{4}$ \\[6pt]
$^{1}$Columbia University \quad
$^{2}$New York University \quad
$^{3}$Microsoft \quad
$^{4}$Peking University \\[3pt]
\texttt{\{sara.ziweigong, tqo2101, lin.ai\}@columbia.edu} \quad \texttt{shawn.yin@nyu.edu}
}

\newcommand{\framework}{E$^2$Sim}

\newcommand{\shawn}[1]{#1}

\definecolor{e2simgreen}{HTML}{5F8F7B}
\definecolor{e2simgreenlight}{HTML}{EEF5F1}
\newcommand{\figstepnum}[1]{#1}

\newtcolorbox{contributionbox}{
    colback=e2simgreenlight,
    colframe=e2simgreen,
    boxrule=0.6pt,
    arc=2pt,
    left=6pt,
    right=6pt,
    top=4pt,
    bottom=4pt,
    before skip=5pt,
    after skip=7pt,
}

\iclrfinalcopy
\begin{document}
\raggedbottom
\maketitle
\lhead{Preprint}
\raggedbottom

\begin{abstract}
AI agents are increasingly expected to support users over long periods, across work, and daily life. This requires modeling not only who users are, but also how their goals, interests, activities, and interactions evolve over time. A challenge in longitudinal user simulation is maintaining consistency between these evolving states and the consequences of actions in a shared environment. We introduce \framework{}, a longitudinal user simulation framework that jointly models evolving actor states and a dynamic shared environment. \framework{} connects actor-state updates with shared-world changes through a feedback loop linking immediate actions, delayed outcomes, and actor-specific observations across work and life. This allows the consequences of earlier actions to shape later behavior, while preserving differences in what each actor knows. We further propose a suite of longitudinal trajectory diagnostics, measuring dynamic consistency, coherent event progression and collaboration, and non-degenerate trajectory development rather than static consistency alone. In one 60-day simulation, \framework{} produced more extracted event threads judged meaningful by our evaluator and fewer near-duplicate actions than four direct-generation baselines. The framework and evaluation methods provide a basis for studying longitudinal user simulation, with potential applications in training and evaluating AI agents that assist users over extended periods.
\end{abstract}

\section{Introduction}
Large language model (LLM) agents are being developed as assistants for everyday work and life tasks \citep{yuan2024personalizedlargelanguagemodel,wu2025longmemevalbenchmarkingchatassistants, meta2026muse, openai2026dots}. A personal assistant that supports the same person over months must reason about multiple projects, stable habits, and evolving goals and interests. Events in one part of a person's life can shape what they do or ask in another, sometimes days or weeks later. An isolated interaction may capture the resulting request while omitting the earlier events that shaped it, making these cross-context effects difficult to observe.

A useful way to study these cross-context effects is to simulate an actor's actions together
with the evolving situation in which those actions occur. Task-level agent benchmarks evaluate the completion of specified objectives \citep{chen2026agentifonedaytasklevelinstructionfollowingbenchmark}, while personalization benchmarks use extended interaction histories to evaluate responses to a user's changing profile \citep{jiang2025knowmerespondme}. These settings do not directly trace how the consequences of everyday actions change an actor's goals and knowledge, and how those changes shape later actions across work and life. Crucially, accumulated history should influence not only \emph{which} action an actor takes next, but also \emph{how} familiar activities are carried out. A PhD student, for example, may spend years reading papers, designing experiments, and writing. Yet the papers they attend to, the experiments they design, the details of their execution, and the outcomes they expect change with accumulated knowledge, goals, and interests. Longitudinal simulation should therefore capture \shawn{how actors perform the same activities differently as their knowledge and goals change.}

This introduces a challenge that is easy to miss when evaluating isolated steps. A trajectory can remain locally plausible while gradually becoming repetitive, returning to the same routines without meaningful progression, or
ceasing to generate consequential new developments.
At the same time, simply \shawn{adding new actions is not enough}: new actions must remain consistent with long-term goals, prior outcomes, and what each actor has actually observed. 
In a multi-actor simulation, an outcome should update an actor's state only when it becomes visible to that actor. For example, an inconclusive experiment may change a researcher's next plan, while a collaborator can respond to the result only after learning about it.
Long-horizon simulation therefore requires \emph{continuity}: later behavior should reflect accumulated history, and \emph{development}: the trajectory should produce \shawn{changes that are consistent with earlier events and affect later behavior, rather than repeat the same routines.}

\shawn{To examine these two requirements, we study the generation of long-horizon everyday action trajectories from an initial scenario that specifies people, their relationships, and their shared world. Even with the full history and explicit instructions to avoid previously observed errors, the language model's outputs do not consistently reflect relevant past information, weakening continuity. Explicitly asking the model to generate realistic and diverse outputs also fails to prevent repetition. We further find that asking it to generate goals and produce actions based on those goals reduces repetition to some extent. However, the resulting behavior is still highly limited by the quality and diversity of the new goals. Goal generation itself also has continuity problems, as the model does not reliably use past information to form high-quality new goals.}

We introduce \framework{} (\emph{Everyday Simulation with Evolving States}), a framework that centers long-horizon simulation on \shawn{the states of simulated character (actors)} with stable attributes and changing goals, plans, and knowledge. At each time slot, \framework{} uses an actor's current state and long-term goals to propose and select work and life goals. \shawn{It then generates candidate action trajectory with different possible outcomes, and applies the realized outcome to both the actor's state and the shared world.} World changes are represented through facts, communication, and delayed follow-up events, and update other actors' states only when they become visible to them. This actor--environment feedback loop allows immediate and delayed consequences to carry across time and actors while preserving differences in what each actor knows.

To evaluate \shawn{\framework{} through the generated} trajectories, we introduce a suite of trajectory-based longitudinal evaluation metrics covering dynamic consistency, event progression, collaboration, and non-degeneracy. In a 60-day comparison with \shawn{four baselines that prompt an LLM to generate trajectories directly}, \framework{} exhibits \shawn{a lower near-duplicate rate and more meaningful events that develop through a series of related actions, with later actions building on earlier outcomes.}

\section{Related Work}
\label{sec:related_work}

\paragraph{Simulation Environments and Frameworks} LLM-based simulation frameworks model agents and their interactions with an environment. Generative Agents combines memory, reflection, and planning to simulate routines and emergent group behavior in a shared sandbox \citep{park2023generative}. Concordia uses a Game Master to resolve agents' actions and determine their effects on the environment \citep{vezhnevets2023generativeagentbasedmodelingactions}. \shawn{Recent updates to Concordia support configurable components for both actors and Game Masters, with actors updating their states and plans from observations \citep{vezhnevets2025multiactorgenerativeartificialintelligence}.} SOTOPIA studies goal-directed social interactions in open-ended scenarios \citep{zhou2024sotopiainteractiveevaluationsocial}, with SOTOPIA-S4 extending the setting to configurable large-scale and multi-party simulation \citep{zhou-etal-2025-sotopia}, and LIFELONG-SOTOPIA carrying interaction histories across multiple social episodes \citep{goel2025lifelong}. Other systems scale LLM-based behavioral simulation to recommendation and social-network environments \citep{wang2025user} or large synthetic societies \citep{piao2026agentsocietylargescalesimulationllmdriven}. Together, these works establish mechanisms for persistent memory, interaction, and environment-mediated simulation.
\begin{contributionbox}
\textbf{Our focus.}
Prior simulation frameworks provide environments and mechanisms for agent interaction and state updates. \framework{} makes the \emph{feedback loop between evolving actor state and shared-world state} the primary object of long-horizon simulation\shawn{, tracing how action outcomes reshape actors' goals and change the course of their ongoing activities.}
\end{contributionbox}


\paragraph{Human Behavior Simulation and Evaluation}
Another line of work evaluates human behavior simulation against reference responses and actions. A study of self-report-grounded agents tests whether agents built from interviews and surveys reproduce the same individuals' responses on surveys and economic games \citep{park2026llmagentsgroundedselfreports}. SimBench evaluates population-level fidelity by comparing model and human response distributions across diverse behavioral tasks \citep{hu2026simbenchbenchmarkingabilitylarge}. BehaviorChain tests whether models can infer successive behaviors from persona information as a scenario unfolds \citep{li-etal-2025-far}. OmniBehavior evaluates predictions on long-horizon, cross-scenario behavior traces from real platform logs \citep{chen2026realworldhumanbehaviorsimulation}. More recent work develops behavior-specialized foundation models \citep{zhou2026odyssim} and evaluates long-horizon activity simulation against continuous real-world activity traces \citep{cheng2026measuring}. These works primarily measure fidelity against observed human responses or future behaviors.
\begin{contributionbox}
\textbf{Our focus.}
Prior behavior-simulation work primarily asks whether generated behavior matches an observed reference. \framework{} asks whether a \emph{generated trajectory remains causally coherent and continues to develop} when no reference future is available.
\end{contributionbox}

\begin{figure}[t]
    \centering
    \includegraphics[width=1\linewidth]{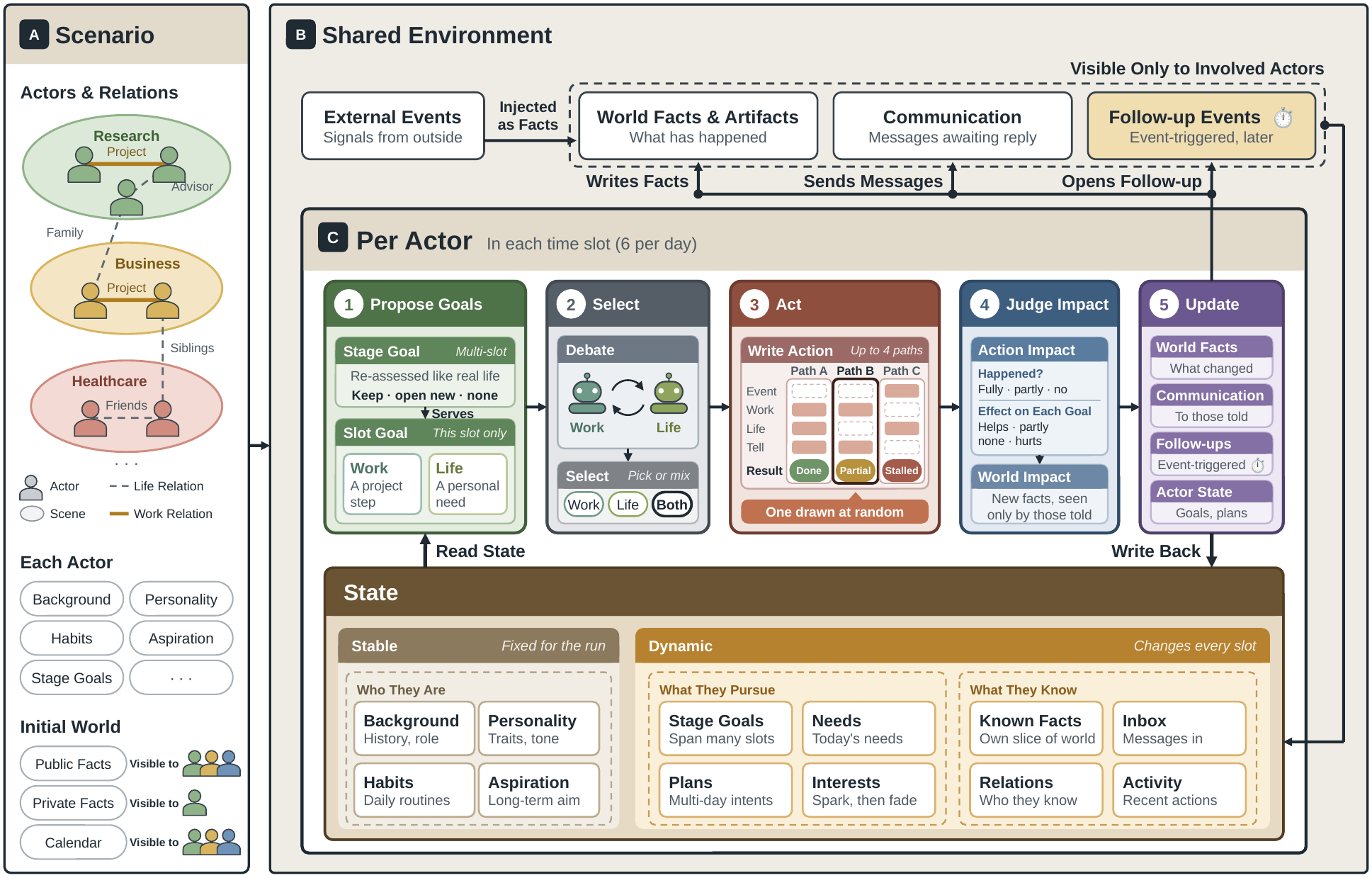}
    \caption{Overview of \framework{}. A scenario (A) initializes actor states and the shared environment (B). At each time slot, actors generate actions from their current states and observations, and the outcomes update both for the next slot (C).}
    \label{fig:overview}
    \par\vspace{-10pt}
\end{figure}

\paragraph{Dynamic User Modeling and Longitudinal Life Simulation}
Related work on longitudinal user modeling studies how representations of an individual change with accumulating experience. PERSONAMEM evaluates whether assistants can infer and track evolving user profiles across repeated interactions \citep{jiang2025knowmerespondme}, while DEEPER continually refines user personas from streaming behavioral observations for future behavior prediction \citep{chen-etal-2025-deeper}. Recent work on dynamic persona coherence similarly separates stable identity from history-dependent adaptive state in sustained role-playing interactions \citep{qi-etal-2026-beyond}, and VitaBench 2.0 evaluates agents that must extract, update, and apply evolving user preferences across temporally ordered tasks \citep{chen2026vitabench}. Most closely related to our setting, LifeSim generates long-horizon life trajectories using a BDI-based cognitive model, physical context, and an event engine, and uses the resulting trajectories to evaluate personalized assistants \citep{duan2026lifesimlonghorizonuserlife}. 
\begin{contributionbox}
\textbf{Our focus.}
Prior longitudinal user models primarily evolve the individual.
\framework{} \emph{jointly evolves the actor and the world}, including shared artifacts,
actor-specific observations, and delayed consequences across multiple actors.
\end{contributionbox}
See Appendix~\ref{app:positioning} for a structured comparison with representative prior work.

\section{Simulation Framework}\label{sec:method}
Existing agent simulation frameworks give agents persistent memory and let them interact in shared environments \citep{park2023generative,vezhnevets2023generativeagentbasedmodelingactions,vezhnevets2025multiactorgenerativeartificialintelligence}. For long-horizon simulation, however, \shawn{we need to track both how actors' goals, plans, and knowledge change and how the consequences of actions persist in the shared world. We therefore jointly model actor states and a shared world state, with actions changing the world and observed world changes updating actor states.}

Figure~\ref{fig:overview} summarizes \framework{}.  \shawn{The scenario (A) initializes actor states and the shared world (B). At each time slot, the per-actor loop (C) uses an actor's current state and world information visible to that actor to select goals and generate actions. The action outcomes then update both the actor's state and the shared world, providing the updated states for the next time slot.}

\subsection{Evolving Actor State}\label{sec:actor_state}
\begin{wrapfigure}{r}{0.6\textwidth}
    \centering
    \vspace{-10pt}
    \includegraphics[width=\linewidth]{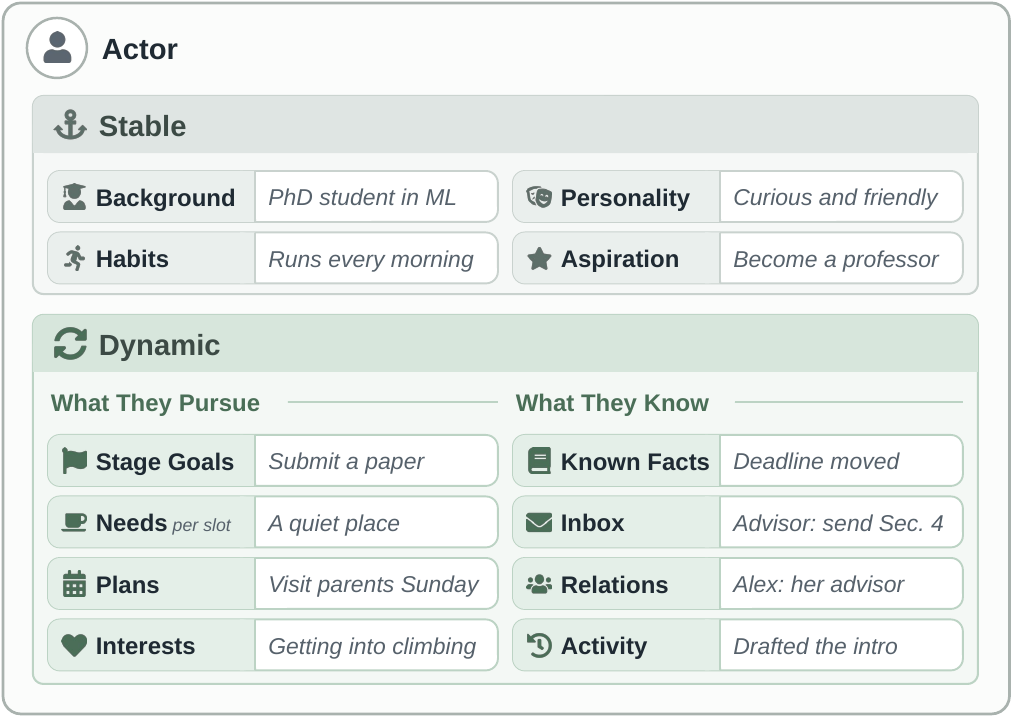}
    \caption{State categories and examples in \framework{}.}
    \label{fig:actor_state}
    \par\vspace{-15pt}
\end{wrapfigure}
In long-horizon simulation, an actor's behavior reflects both enduring characteristics and a changing situation. We therefore represent the state of actor $i$ at time $t$ as $S_i^t$, with stable and dynamic components (Figure~\ref{fig:actor_state}). Figure~\ref{fig:overview}A shows how the scenario initializes both components: actor profiles supply stable attributes, while initial \emph{Stage Goals} and actor-visible world facts initialize parts of the dynamic state. The dynamic state then evolves across time slots.

The stable state captures attributes that distinguish actors and remain fixed within a simulation run. \emph{Background} describes an actor's history and role. \emph{Personality} describes traits that affect how the actor responds to different situations. \emph{Habits} provide recurring routines, and \emph{Aspiration} gives the actor a long-term direction. For example, a PhD student who prefers to work alone, runs each morning, and aspires to become a professor may go for a morning run and then revise a paper draft independently after receiving advisor comments. An actor with the same background, habits, and aspiration but a personality that favors discussion may first meet the advisor to clarify the comments before revising.

The dynamic state captures the actor's changing objectives and its experience of events and interactions during the simulation. \emph{Stage Goals} specify outcomes the actor works toward over multiple time slots, such as submitting a paper. \shawn{Each actor maintains a set of active Stage Goals. At each time slot, it uses its long-term aspiration, current state, and observations to select an existing goal, propose a new one, or select none. Unselected goals remain active and retain their progress.} Completing a Stage Goal requires evidence for each specified completion condition. \shawn{The model identifies whose records can establish that condition, including those of another actor or an external system when their confirmation is required.} \emph{Plans} record particular activities it intends to carry out, such as revising a section tomorrow. \emph{Interests} record topics or activities that attract the actor's attention and may later fade. \emph{Needs} represent immediate work or life concerns generated for each slot. The state also contains \emph{Known Facts} the actor has learned about the environment, messages received in the \emph{Inbox}, \emph{Relations} with other actors, and \emph{Activity} recording its recent actions. \shawn{We retain a history of updates to Stage Goals, plans, and interests and reconstruct their current states from this history.}

\shawn{Action outcomes and observed events can change what an actor should do next, and these effects may last across multiple time slots. We therefore update the dynamic state to retain these effects for later decisions.} The action's action outcome may advance a \emph{Stage Goal}, change a \emph{Plan}, while the action itself is recorded in \emph{Activity}. Observed world changes may update \emph{Known Facts} or add messages to the \emph{Inbox}, or lead the actor to revise its goals or plans.

\subsection{Actor and Environment Dynamics}
\label{sec:actor_environment}
\shawn{Having defined the actor state, we now describe how actors and the shared environment affect each other. An actor uses its state and current observation to propose an action, and the simulator then determines its outcome and updates the shared world accordingly. In turn, world changes that become visible to an actor update its dynamic state and shape its later actions. Because actors differ in their states and observations, the same world change may lead to different responses.}

\subsubsection{Actions Change the Environment}
\label{sec:action_effects}
\shawn{Actions generation} begins by proposing and selecting slot goals based on the actor's current state (Steps~\mbox{\figstepnum{1}--\figstepnum{2}} in Figure~\ref{fig:overview}). We derive slot goals that specify \shawn{tasks suited to the time available} for the current slot because, in preliminary simulations, actors acting directly on multi-day \emph{Stage Goals} compressed months-long research projects into days. \shawn{We also observed that a model given a work slot goal and prompted to consider life needs produced unrealistic life actions in every slot, with greater emphasis on those life needs than on the work goal. We therefore use separate work and life proposers to generate slot goals from the actor's state, including its personality. The proposers exchange their proposals and rationales to debate which goals should take priority in the current situation. A selector then chooses the work goal, the life goal, or both for the current time slot.}

\shawn{Based on this goal selection}, the actor generates several candidate action paths with different possible outcomes, \shawn{from positive to negative}, and samples one uniformly (Step~\figstepnum{3} in Figure~\ref{fig:overview}). \shawn{In preliminary runs, we observed that the actor tended to select actions with positive outcomes even when prompted to consider setbacks, and a separate model used for path selection showed the same preference. This often led to unrealistically fast completion of long-term tasks that normally require multiple rounds of feedback and revision. We found that generating multiple candidate actions and selecting one at random reduced this bias effectively so we include this randomness in the framework. Moreover, an actor's limited view of the scenario may lead it to propose actions that fit its goals but may not occur in real life. The selected path therefore passes to a judge (Step~\figstepnum{4} in Figure~\ref{fig:overview}), which determines whether the action can occur fully, partly, or not at all under the current conditions. The judge then assesses the effects of the realized action on the actor's \emph{Stage Goals} and the shared world.}

The simulator then applies the judged outcome (Figure~\ref{fig:overview}, Step~\figstepnum{5}). Depending on what occurred, the update can revise world facts and affected actors' dynamic states, send messages to relevant actors, and schedule follow-up events for consequences that arise later. For example, an agreement today to travel next week can update the actors' plans now and schedule a follow-up event for the trip.

\subsubsection{Environment Changes Shape Actions}
\label{sec:environment_feedback}
The environment influences later actions through each actor's dynamic state. We distinguish two sources of environmental change: the effects of earlier actor actions and external events introduced independently of those actions. In both cases, actors update their states from the changes they observe, and the updated states guide their next actions.

\shawn{World facts record changes to shared objects and conditions, communication conveys information to the actors involved, and follow-up events introduce delayed effects at scheduled times (Step~\figstepnum{5} in Figure~\ref{fig:overview}). As these effects become visible to an actor, they update the relevant parts of its dynamic state, such as known facts, goals, or plans.}

\shawn{External events can change an actor's priorities by introducing new opportunities or constraints}. We therefore introduce an external event component that \shawn{generates events conditioned on the current scenario, with randomness in both their content and timing. Events are made visible either to selected actors or to all actors, and each recipient decides whether to act on them.} Each actor assesses \shawn{the event based on its own state and may update its knowledge, revise its plans, or develop a new \emph{Interest}}. For example, after a concert is announced, an actor with a background in music and art may add the concert to its \emph{Interests}, seek more information later, and eventually decide to attend.

\subsection{Multi-Actor Interaction}
\label{sec:multi_actor}
The preceding sections describe how an actor's actions change the environment and, through observation, its own state. Everyday activities also involve other actors who may affect the same project or situation without having the same information. \framework{} therefore maintains one shared environment \shawn{while giving each actor its own state and observations}. An actor's action can affect another actor when its outcome becomes visible to that \shawn{actor's later behavior when its outcome becomes visible to that actor} and updates its dynamic state.

Shared projects and artifacts allow the results of different actors' actions to accumulate in the environment. For example, when one actor revises a paper, the revision remains in the shared document, allowing a collaborator to adjust its plan once it observes the change.

Messages pass through the environment and enter each intended recipient's \emph{Inbox} when delivered. Collaboration or conflict may also change an actor's \emph{Relations}, which record its own view of other actors. Participants can therefore differ in what they know and how they view one another.

When several actors act in the same time slot, each generates its action from \shawn{its own state and observations of a shared environment snapshot frozen at the start of the slot. After all actors finish, the simulator checks concurrent completion claims for shared goals against their completion conditions and supporting evidence. A confirmed completion closes the associated \emph{Stage Goals} for all actors sharing the project. The simulator then merges the updates and makes the resulting changes available for observation in the next slot. This prevents an actor from responding to another actor's action simply because it was generated earlier in the same slot.}

\subsection{Simulation Workflow and Implementation}
\label{sec:simulation_loop}
\begin{figure}[t]
    \centering
    \includegraphics[width=1\linewidth]{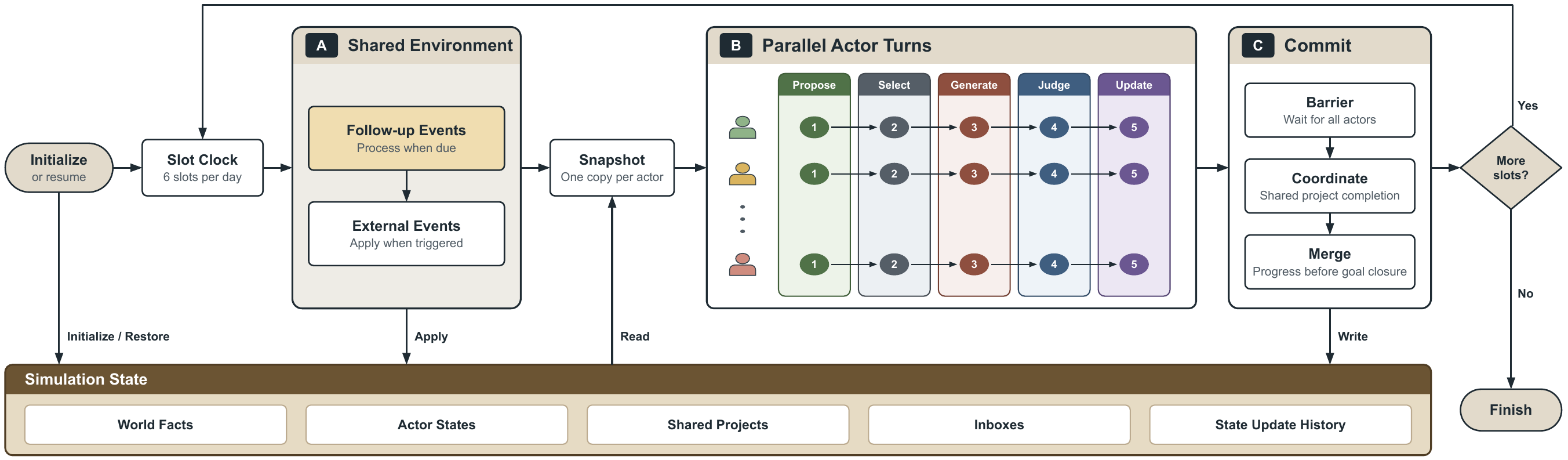}
    \caption{Execution workflow of \framework{}. Each slot updates the shared environment (A), runs actor turns in parallel from a frozen snapshot (B), and merges their updates after all actors finish (C).}
    \label{fig:system_overview}
    \par\vspace{-10pt}
\end{figure}
\shawn{\framework{} initializes actor states and a shared environment from the scenario, then advances the simulation through successive time slots. Each slot begins by processing due follow-up events and any new external events. Actors observe available changes and update their dynamic states before generating actions from a frozen environment snapshot. Action generation and outcome judging run in parallel across actors. After all actors finish, the simulator merges their updates to actor and world states, making the resulting changes available for observation in the next slot.}

\shawn{Figure~\ref{fig:system_overview} shows the execution architecture. A Python runner manages the slot calendar and shared simulation state. The same component functions are also registered in a LangGraph reference graph used for single-actor probes. After the environment phase (A), the runner freezes the state and dispatches actor turns to a thread pool, with each turn operating on an isolated copy. Actors run in parallel, while component calls within each turn remain sequential (B). Once all turns finish, the runner coordinates shared-project completion and merges their updates into the shared simulation state, recording progress before applying goal closures (C). Additional optimizations to model calls, inputs, and state management are detailed in Appendix~\ref{app:system_optimization}.}

\section{Longitudinal Evaluation}
\label{sec:eval}
To assess whether trajectories remain coherent and continue to develop, we introduce four groups of longitudinal metrics for dynamic consistency, event progression, collaboration, and non-degeneracy. A LLM-based evaluator scores eligible actor transitions, event threads, and collaboration episodes. These scores characterize coherence and experiment within the simulated world, not fidelity to real individuals or effectiveness in downstream assistant tasks.

\paragraph{Individual Dynamic Consistency (IDC).}
IDC asks whether changes in communication, behavior, goals, and interests follow from an actor's prior experience. It also identifies unsupported changes, missed adaptations, and spillover to unrelated facets. IDC Macro averages alignment over eligible transitions within actors and then across actors on a $0$--$100$ scale. Higher values indicate stronger support from history. Lower Missed Adaptation Rate and Spillover Rate indicate fewer corresponding errors.

\paragraph{Emergent Event Evaluation (EEE).}
With no predefined event list, we extract action threads around a goal, problem, or shared activity. EEE asks whether each thread has a grounded start (Event Initiation Plausibility, EIP), progresses consistently with earlier outcomes and elapsed time (Event Progression Coherence, EPC), and has a status supported by its history (Event Status Appropriateness, ESA). A thread may properly remain open or waiting. All three scores use a $0$--$100$ scale over eligible threads, with higher values indicating stronger support.

\paragraph{Event Collaboration Dynamics (ECD).}
A shared event may progress coherently overall while a participant forgets a commitment or acts before a dependency is met. ECD evaluates each decision against the participant's role, commitments, event state, and other actors' steps. Acting and waiting may both be appropriate. Episode Collaboration Plausibility (ECP) summarizes coordination on a $0$--$100$ scale, with higher scores indicating better coordination. Context-Sensitive Response Calibration (CSRC) checks whether responses vary appropriately across situations. ECD measures coordination within shared events, not general relationship quality.

\paragraph{Non-degeneracy Diagnostics (NDD).}
High scores on eligible transitions can coexist with inactive or repetitive trajectories. NDD counts meaningful event threads per $1{,}000$ actor-slots (one actor in one time slot) and measures near-duplicate actions. It also checks whether new threads are grounded in context, events have later consequences, and decisions use relevant history. Lower duplication is preferable, whereas event density has no universal target and must be read with the other diagnostics.

\paragraph{Evaluation reliability.}
A post-hoc judge may reward an observed response based on a plausible explanation constructed afterward. We require trajectory evidence and test the judge on injected errors, including forgotten commitments and violated dependencies. Perturbation Win Rate (PWR) is the share of original--corrupted episode pairs for which the judge scores the original higher. Higher PWR indicates better detection of these errors, not agreement with human ratings. Appendix~\ref{app:eval_impl} details the rubrics, extraction procedure, and aggregation rules.

\section{Experiment}
\label{sec:experiment}
We generate 60-day action trajectories for ten actors within one scenario spanning four connected settings, with six time slots per day. Each trajectory contains 3,600 actor-slot action records. We run \framework{} and each of four baselines once from the same scenario file. All five generation runs use \texttt{gpt-5.6-luna}. Appendix~\ref{app:scenario_init} details the actor roster, initial goals, and world configuration.

We compare \framework{} with four \shawn{baselines that generate action trajectories directly from the supplied context using LLM}. \textbf{Direct} prompts the model once per day with the full scenario and the trajectory generated so far to generate that day's actions for all actors. \textbf{Guided} retains this setup but adds instructions targeting errors observed in Direct, testing how far prompting alone can address them. \textbf{Per-Actor} generates each actor's actions separately using its own profile, visible initial facts, its own history, and other actors' entries that mention it. \textbf{Summary} retains joint daily generation but replaces the full history with a summary maintained by the model and \shawn{action records from the most recent three days}, examining the effect of a more compact history. We evaluate the five generated trajectories using the longitudinal measures in Section~\ref{sec:eval}. \shawn{We additionally compare \framework{} with native Concordia \citep{vezhnevets2023generativeagentbasedmodelingactions}. Starting from the same source scenario, we run Concordia for seven simulated days and compare it with the first seven days of \framework{}'s trajectory, covering 42 time slots and 420 scheduled actor actions per system. We compare estimated API cost and generation time using the same accounting protocol (Appendix~\ref{app:eval}) and report qualitative observations from the generated actions and outcomes.}

We use \texttt{gpt-5.6-luna} with medium reasoning effort as the evaluation judge. \shawn{Cases with insufficient evidence are excluded from the relevant metric. Evaluation coverage is reported in Appendix~\ref{app:eval_coverage}.} The extraction and scoring procedures are detailed in Appendix~\ref{app:eval_impl}.

\subsection{Baseline Comparison}
\begin{table*}[!t]
\caption{Longitudinal results from one 60-day run per method. Event quality scores cover eligible extracted threads or episodes. \shawn{See Section~\ref{sec:eval} for metric definitions. In particular, the Collaboration Plausibility row reports Episode Collaboration Plausibility.} Bold marks the best value per metric.}
\label{tab:baseline_comparison}
\centering
\resizebox{\textwidth}{!}{
\begin{tabular}{@{}c|l|c|ccccc@{}}
\toprule
\textbf{Group} & \textbf{Metric} & \textbf{Scale}
& \cellcolor{violet!8}\textbf{\framework}
& \textbf{Direct}
& \textbf{Guided}
& \textbf{Per-Actor}
& \textbf{Summary} \\
\midrule
\multirow{3}{*}{\textbf{IDC}}
& IDC Macro $\uparrow$ & 0--100
& \cellcolor{violet!8}52.06 & \textbf{56.51} & 52.87 & 54.24 & 47.75 \\
& Missed Adaptation Rate $\downarrow$ & \%
& \cellcolor{violet!8}\textbf{27.4} & 33.4 & 35.6 & 29.1 & 33.9 \\
& Spillover Rate $\downarrow$ & \%
& \cellcolor{violet!8}54.0 & 47.5 & 48.9 & \textbf{47.2} & 56.2 \\
\midrule
\textbf{ECD}
& Collaboration Plausibility $\uparrow$ & 0--100
& \cellcolor{violet!8}93.25 & \textbf{95.92} & 93.00 & 95.24 & 92.76 \\
\midrule
\multirow{3}{*}{\textbf{EEE}}
& Event Initiation Plausibility $\uparrow$ & 0--100
& \cellcolor{violet!8}\textbf{87.74} & 77.34 & 83.06 & 85.94 & 85.34 \\
& Event Progression Coherence $\uparrow$ & 0--100
& \cellcolor{violet!8}88.86 & 71.33 & 84.58 & \textbf{90.10} & 84.57 \\
& Event Status Appropriateness $\uparrow$ & 0--100
& \cellcolor{violet!8}\textbf{94.85} & 82.14 & 92.19 & 93.06 & 94.23 \\
\midrule
\multirow{2}{*}{\textbf{NDD}}
& Near-Duplicate Actions $\downarrow$ & \%
& \cellcolor{violet!8}\textbf{22.3} & 73.5 & 62.8 & 60.5 & 71.0 \\
& Meaningful Event Threads $\uparrow$ & per 1k actor-slots
& \cellcolor{violet!8}\textbf{21.4} & 3.9 & 10.3 & 8.3 & 6.4 \\
\bottomrule
\end{tabular}}
\end{table*}

\paragraph{Prompting Baselines.} Table~\ref{tab:baseline_comparison} compares \shawn{\framework{} with the four prompting baselines over 60 simulated days.} \framework{} has a higher density of extracted threads judged meaningful by our evaluator than any baseline (21.4 versus 3.9--10.3 per 1,000 actor-slots) and a lower near-duplicate rate (22.3\% versus 60.5--73.5\%). It also receives the highest event initiation and status scores, while Per-Actor scores slightly higher on event progression. The consistency results are mixed: \shawn{\framework{} has the lowest missed adaptation rate, but} Direct has higher IDC and collaboration plausibility, and three baselines have lower spillover rates. \shawn{Overall, the clearest observed differences concern the density of meaningful event threads and repetition, while consistency results remain mixed.}

\shawn{The thread extractor links 1,276 of \framework{}'s 3,600 action records to event threads, compared with 1,906--2,298 records for the baselines}. The event quality scores therefore describe different eligible subsets (Appendix~\ref{app:eval_coverage}), while meaningful-thread density also depends on how the extractor groups actions. Each method was run once, so the comparison describes the observed trajectories without estimating variation across independent runs.

\begin{figure}[H]
    \centering
    \begin{minipage}[c]{0.70\linewidth}
        \centering
        \IfFileExists{figs/e2sim_concordia_qualitative.pdf}{%
            \includegraphics[width=\linewidth]{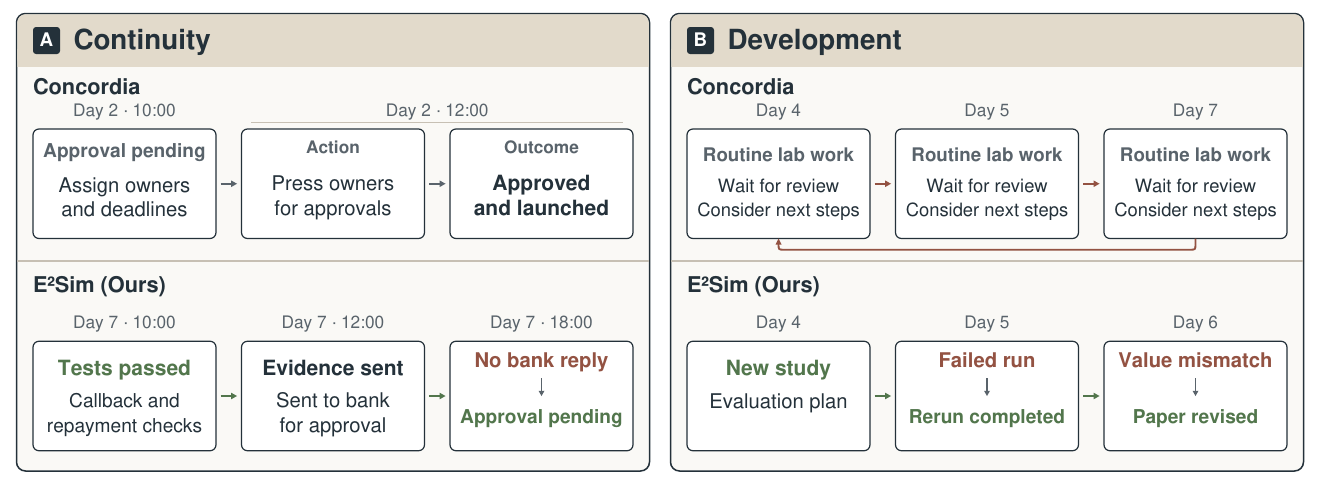}%
        }{%
            \IfFileExists{e2sim_concordia_qualitative.pdf}{%
                \includegraphics[width=\linewidth]{e2sim_concordia_qualitative.pdf}%
            }{%
                \fbox{\parbox[c][0.35\linewidth][c]{\dimexpr\linewidth-2\fboxsep-2\fboxrule\relax}{%
                    \centering\footnotesize Figure awaiting upload\\[4pt]
                    \texttt{e2sim\_concordia\_qualitative.pdf}}}%
            }%
        }
        \par\vspace{-10pt}
        \caption{Continuity and development in selected trajectory excerpts.}
        \label{fig:concordia_comparison}
    \end{minipage}\hfill
    \begin{minipage}[c]{0.29\linewidth}
        \centering
        \footnotesize
        \setlength{\tabcolsep}{3pt}
        \renewcommand{\arraystretch}{1.15}
        \begin{tabular}{@{}lrr@{}}
            \toprule
            & \shortstack{Est. cost\\(USD)} & \shortstack{Time\\(min)} \\
            \midrule
            Concordia & 102.69 & 605.54 \\
            \framework{} & \textbf{6.83} & \textbf{71.25} \\
            \midrule
            Reduction & \textbf{93.3\%} & \textbf{88.2\%} \\
            \bottomrule
        \end{tabular}
        \captionsetup{justification=raggedright,singlelinecheck=false}
        \captionof{table}{Estimated API cost and generation time over seven simulated days.}
        \label{tab:concordia_efficiency}
    \end{minipage}
    \par\vspace{-10pt}
\end{figure}

\paragraph{Concordia Baseline.}
\shawn{We directly compare \framework{} with Concordia, a closely related simulation framework, through seven-day simulations starting from the same initial scenario. We use Concordia's native agent and game-master components. In this comparison, some Concordia trajectories compress the resolution of multiple pending requirements into a single outcome or repeatedly describe ongoing activities without concrete changes across days. Figure~\ref{fig:concordia_comparison} illustrates these patterns through selected examples of continuity and development.}

\shawn{The continuity example concerns a product launch that requires external approvals. These approvals must be confirmed before the launch is considered complete. Concordia reports obtaining the required approvals and launching the product together, leaving the intermediate steps unspecified in the selected record. \framework{} separately tracks successful testing, the approval request, and the bank's response, keeping approval pending when no response has arrived. The development example concerns whether recurring research activities lead to concrete new steps. Across the selected days, Concordia repeatedly describes routine lab work, waiting for review, and considering future plans. In contrast, \framework{} follows a new study through a failed run, a rerun, and revision following feedback, with earlier outcomes shaping subsequent actions.}

\shawn{Meanwhile, Table~\ref{tab:concordia_efficiency} shows that \framework{} reduces estimated API cost from \$102.69 to \$6.83 and generation time from 605.54 to 71.25 minutes (see Appendix~\ref{app:eval} for the cost estimation procedures).}

\subsection{Long-Horizon Trajectory Dynamics}
\label{sec:long_horizon_dynamics}
\begin{figure}[t]
    \centering
    \includegraphics[width=0.87\linewidth]{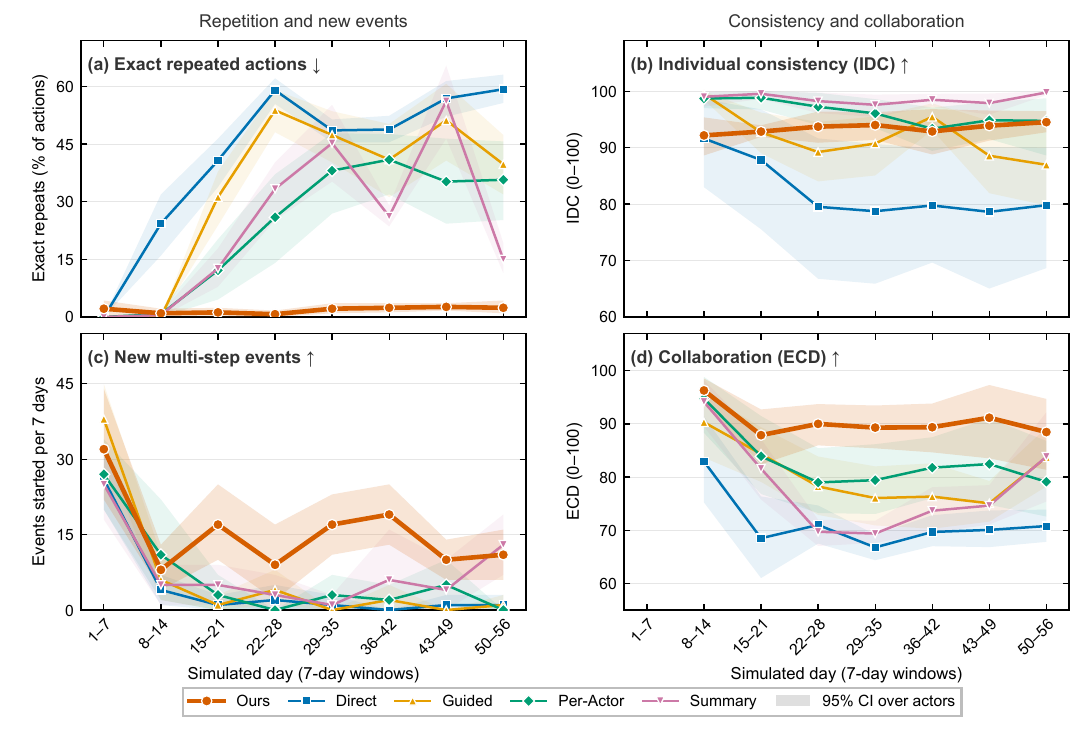}
    \caption{Trajectory measures across days 1--56. Shading shows 95\% bootstrap intervals over actors. Exact repeats in (a) differ from judged near-duplicates in Table~\ref{tab:baseline_comparison}.}
    \label{fig:stability}
    \par\vspace{-10pt}
\end{figure}

The aggregate comparison does not show how repetition and event activity change over time. We use the first 56 days of each run to form eight nonoverlapping seven-day windows and track exact action repetition, new multi-step events, individual dynamic consistency (IDC), and unsupported drift (Figure~\ref{fig:stability}). \shawn{We compute 95\% bootstrap intervals by resampling the ten actors. These intervals do not capture variation across independent simulation runs.}

\shawn{Exact repetition stays around 1--3\% for \framework{}, while most baselines reach roughly 35--60\% in later windows (Figure~\ref{fig:stability}a).} The estimated increase is 0.18 percentage points per window for \framework{}, versus 5.19--7.05 for the baselines. \shawn{The gap remains large in later windows, although baseline repetition rates fluctuate.} The event-start series suggests a related pattern (Figure~\ref{fig:stability}c). \shawn{All systems start many new multi-step events in the first window, but their numbers remain higher for \framework{} than for most baselines in later windows (Figure~\ref{fig:stability}c). Unsupported drift rates stay within a similar range across methods (Figure~\ref{fig:stability}d), while Direct maintains higher IDC than \framework{} (Figure~\ref{fig:stability}b). Together, these trends show that \framework{} sustains event activity while keeping exact repetition low.}

\subsection{Component Analysis}
\label{sec:component_analysis}

\begin{figure}[t]
    \centering
    \includegraphics[width=0.87\linewidth]{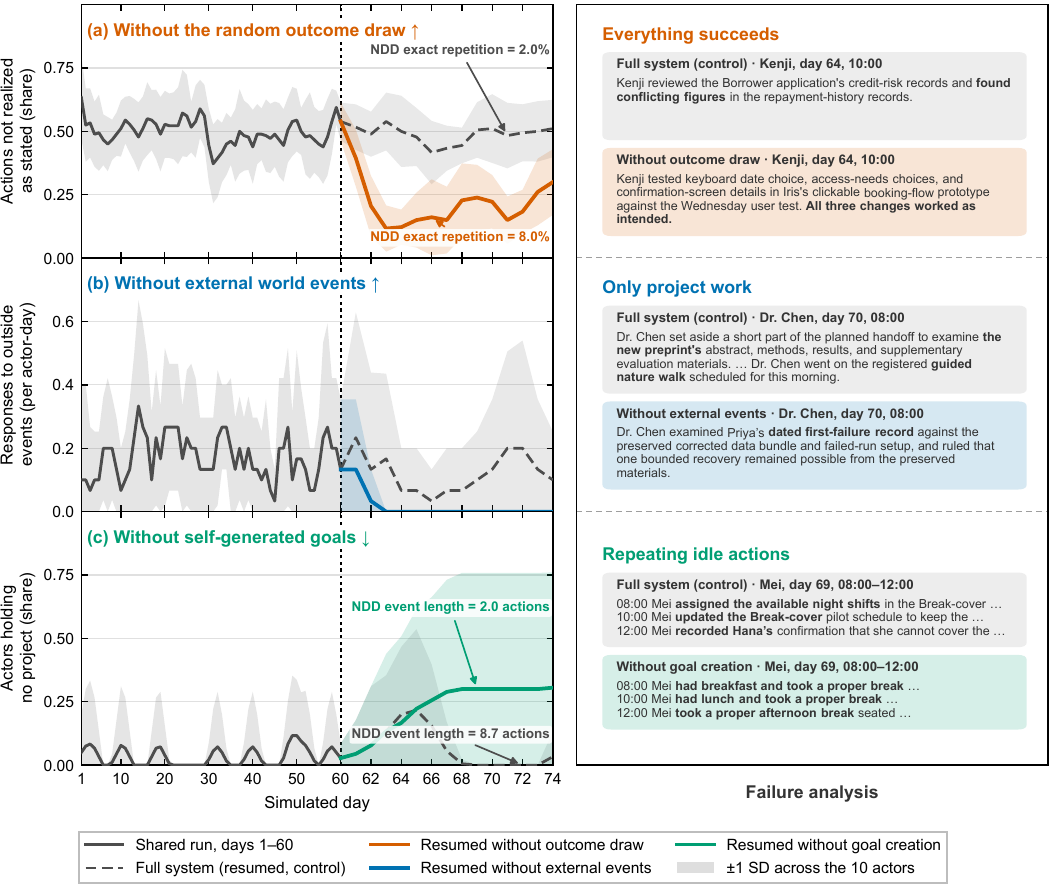}
    \caption{Ablations over 14 days from the same day-60 state. Shading shows one standard deviation across ten actors. Examples show the full system and each ablation.}
    \label{fig:ablation}
    \par\vspace{-10pt}
\end{figure}

From the day-60 checkpoint, we run one 14-day continuation for the full system and each ablation. The three ablations separately remove the random selection among candidate paths, external events, or new \emph{Stage Goal} creation. The conditions use the same model and otherwise identical code, with switches verified in invocation logs. Figure~\ref{fig:ablation} shows metrics trends and ablation failure examples.

Without the random selection, the share of actions not realized as stated falls from 49\% to 19\%, while exact repetition rises from 2.0\% to 8.0\% (Figure~\ref{fig:ablation}a). In matched examples, all three intended changes succeed without the draw, whereas the full system encounters conflicting records. This is consistent with the model's preference for favorable outcomes observed in preliminary runs.

Without external events, recorded responses to them fall to zero by construction (Figure~\ref{fig:ablation}b). In the matched examples, the full system responds to a new preprint and a planned walk, while the ablation continues project work. These examples show how external events can redirect ongoing activities, an effect our current evaluation metrics do not directly measure.

Without new \emph{Stage Goals} creation, 30\% of actor-slots lack an active project on days 68--74, versus 1\% in the full system (Figure~\ref{fig:ablation}c). This delayed increase suggests that existing goals initially sustain project activity. Average new-thread length falls from 8.7 to 2.0 actions. In the matched example, Mei shifts from break-cover coordination to repeated breaks. IDC nevertheless rises from 53.6 to 58.5, showing that higher consistency score does not necessarily indicate sustained development.

For each ablation, we obtained two LLM judgments comparing the 14-day traces under randomized A/B labels. All six judgments correctly identified the ablated trace. The evaluators were told which component had been removed, so this check assesses whether its effects are recognizable in the traces, not whether the full-system traces are more realistic.

\section{Conclusion}
We introduced \framework{}, which preserves stable actor characteristics while updating goals, plans, and knowledge through interactions with a shared world. We also proposed longitudinal measures of state consistency, event development, collaboration, and repetition. In one 60-day run, \framework{} produced a higher density of extracted threads judged meaningful by our evaluator and fewer near-duplicate actions than four prompting baselines. Component analyses show how random selection, external events, and new goals affect long-term activity.

\clearpage
\subsection*{AI use statement}
In this work, we used generative AI tools as core components of the proposed simulation framework, the baseline systems, and the model-based evaluation pipeline, as described in the main text and appendix. We also used generative AI tools for polishing the writing of the paper. We have reviewed all AI-assisted work. AI-assisted code and generated outputs were manually inspected and validated where applicable. We take responsibility for the final content of this work,
including text, claims or artifacts produced with the aid of generative AI.


\subsection*{Ethics statement}

This work studies simulated longitudinal human behavior and does not involve human subjects or private personal data. The generated trajectories are synthetic and are intended for research on long-horizon user simulation and evaluation.

Because such simulations may appear behaviorally plausible, they should not be interpreted as faithful representations of real individuals or used to make high-stakes decisions about people. In particular, our evaluation measures coherence within the simulated world rather than fidelity to real human behavior. Future applications of longitudinal user simulation should therefore consider risks related to overgeneralization, stereotyping, privacy, and inappropriate reliance on simulated behavior in sensitive settings.

\subsection*{Reproducibility statement}

We provide the information needed to reproduce the simulation and evaluation protocol across the main text and appendices. Section~\ref{sec:method} specifies \framework{}'s actor-state representation, actor--environment transition mechanism, multi-actor visibility, and simulation loop, while Section~\ref{sec:eval} defines the longitudinal evaluation metrics and their interpretation. Section~\ref{sec:experiment} reports the experimental setup, including the 60-day, ten-actor scenario, baseline definitions, generation model, and evaluation model. Appendix~\ref{app:eval_impl} documents the evaluation pipeline, formal metric definitions, aggregation procedures, eligibility criteria, and abstention rules. Appendix~\ref{app:sotopia_adapter} describes the adapted SOTOPIA evaluator. Appendix~\ref{app:scenario_init} specifies the actor roster, initial goals, relationships, and world configuration used across conditions.

We additionally report metric coverage and denominators for model-based judgments, and explicitly note that each main comparison condition is represented by a single trajectory, so the reported results should not be interpreted as estimates of run-to-run variability.


\subsubsection*{Acknowledgments}
This work is in part supported by the funds provided by the National Science Foundation and by
DoD OUSD (R\&E) under Cooperative Agreement
PHY-2229929 (The NSF AI Institute for Artificial
and Natural Intelligence). The views, opinions
and/or findings expressed are those of the authors
and should not be interpreted as representing the
official views or policies of the National Science
Foundation or the U.S. Government.

\bibliography{iclr2027_conference}

\begin{thebibliography}{24}
\providecommand{\natexlab}[1]{#1}
\providecommand{\url}[1]{\texttt{#1}}
\expandafter\ifx\csname urlstyle\endcsname\relax
  \providecommand{\doi}[1]{doi: #1}\else
  \providecommand{\doi}{doi: \begingroup \urlstyle{rm}\Url}\fi

\bibitem[Chen et~al.(2025)Chen, Du, Chen, Xu, Zhang, Yuan, Chen, Li, and Xiao]{chen-etal-2025-deeper}
Aili Chen, Chengyu Du, Jiangjie Chen, Jinghan Xu, Yikai Zhang, Siyu Yuan, Zulong Chen, Liangyue Li, and Yanghua Xiao.
\newblock {DEEPER} insight into your user: Directed persona refinement for dynamic persona modeling.
\newblock In Wanxiang Che, Joyce Nabende, Ekaterina Shutova, and Mohammad~Taher Pilehvar (eds.), \emph{Proceedings of the 63rd Annual Meeting of the Association for Computational Linguistics (Volume 1: Long Papers)}, pp.\  24157--24180, Vienna, Austria, July 2025. Association for Computational Linguistics.
\newblock ISBN 979-8-89176-251-0.
\newblock \doi{10.18653/v1/2025.acl-long.1177}.
\newblock URL \url{https://aclanthology.org/2025.acl-long.1177/}.

\bibitem[Chen et~al.(2026{\natexlab{a}})Chen, Xu, Cao, Pan, Zhang, Hu, Du, Gao, Lu, Sun, Han, Sun, Wu, and Lin]{chen2026realworldhumanbehaviorsimulation}
Jiawei Chen, Ruoxi Xu, Boxi Cao, Ruotong Pan, Yunfei Zhang, Yifei Hu, Yong Du, Tingting Gao, Yaojie Lu, Yingfei Sun, Xianpei Han, Le~Sun, Xiangyu Wu, and Hongyu Lin.
\newblock Towards real-world human behavior simulation: Benchmarking large language models on long-horizon, cross-scenario, heterogeneous behavior traces, 2026{\natexlab{a}}.
\newblock URL \url{https://arxiv.org/abs/2604.08362}.

\bibitem[Chen et~al.(2026{\natexlab{b}})Chen, Wu, Hou, Tang, Hu, Hou, Li, Qian, Wang, Chen, Tian, Zhang, Bian, Pan, Zhang, Zhou, Cai, Rao, Ren, Huang, Huang, Yuan, Guo, Tang, Zhang, Chen, Chen, Li, Guo, Zuo, Guo, Wang, Yu, Wang, Jiang, Tian, Zhang, Liu, Zeng, Shan, Yin, Hu, Liu, Ren, and Gong]{chen2026agentifonedaytasklevelinstructionfollowingbenchmark}
Kaiyuan Chen, Qimin Wu, Taiyu Hou, Tianhao Tang, Xueyu Hu, Yuchen Hou, Bikun Li, Chengming Qian, Guoyin Wang, Haolin Chen, Haotong Tian, Haoye Zhang, Haoyu Bian, Hongbing Pan, Hongkang Zhang, Hongyi Zhou, Jiaqi Cai, Jiewu Rao, Jiyuan Ren, Keduan Huang, Lucia~Zhu Huang, Mingyu Yuan, Naixu Guo, Qicheng Tang, Qinyan Zhang, Shuai Chen, Siheng Chen, Ting~Ting Li, Xiaoxing Guo, Yaocheng Zuo, Yaoqi Guo, Yinan Wang, Yinzhou Yu, Yize Wang, Yuan Jiang, Yuan Tian, Yuanshuo Zhang, Yuxuan Liu, Yvette~Yan Zeng, Zenyu Shan, Zihan Yin, Xiaobo Hu, Yang Liu, Yixin Ren, and Yuan Gong.
\newblock Agentif-oneday: A task-level instruction-following benchmark for general ai agents in daily scenarios, 2026{\natexlab{b}}.
\newblock URL \url{https://arxiv.org/abs/2601.20613}.

\bibitem[Chen et~al.(2026{\natexlab{c}})Chen, Zhang, Cai, Shi, Yao, Cui, Zheng, Huo, Su, Gu, et~al.]{chen2026vitabench}
Yuxin Chen, Yi~Zhang, Zhengzhou Cai, Yaorui Shi, Zhiyuan Yao, Chenhang Cui, Jingnan Zheng, Yaqi Huo, Xi~Su, Qi~Gu, et~al.
\newblock Vitabench 2.0: Evaluating personalized and proactive agents in long-term user interactions.
\newblock \emph{arXiv preprint arXiv:2605.27141}, 2026{\natexlab{c}}.

\bibitem[Cheng et~al.(2026)Cheng, Yang, Bas, Niinuma, Abe, and Lindlbauer]{cheng2026measuring}
Yi~Fei Cheng, Fan Yang, Iremsu Bas, Koichiro Niinuma, Narishige Abe, and David Lindlbauer.
\newblock Measuring the behavioral fidelity of long-horizon human activity simulations, 2026.
\newblock URL \url{https://arxiv.org/abs/2609.01257}.

\bibitem[Duan et~al.(2026)Duan, Huang, and Wei]{duan2026lifesimlonghorizonuserlife}
Feiyu Duan, Xuanjing Huang, and Zhongyu Wei.
\newblock Lifesim: Long-horizon user life simulator for personalized assistant evaluation, 2026.
\newblock URL \url{https://arxiv.org/abs/2603.12152}.

\bibitem[Goel \& Zhu(2025)Goel and Zhu]{goel2025lifelong}
Hitesh Goel and Hao Zhu.
\newblock Lifelong sotopia: Evaluating social intelligence of language agents over lifelong social interactions.
\newblock \emph{arXiv preprint arXiv:2506.12666}, 2025.

\bibitem[Hu et~al.(2026)Hu, Baumann, Lupo, Collier, Hovy, and Röttger]{hu2026simbenchbenchmarkingabilitylarge}
Tiancheng Hu, Joachim Baumann, Lorenzo Lupo, Nigel Collier, Dirk Hovy, and Paul Röttger.
\newblock Simbench: Benchmarking the ability of large language models to simulate human behaviors, 2026.
\newblock URL \url{https://arxiv.org/abs/2510.17516}.

\bibitem[Jiang et~al.(2025)Jiang, Hao, Cho, Li, Yuan, Chen, Ungar, Taylor, and Roth]{jiang2025knowmerespondme}
Bowen Jiang, Zhuoqun Hao, Young-Min Cho, Bryan Li, Yuan Yuan, Sihao Chen, Lyle Ungar, Camillo~J. Taylor, and Dan Roth.
\newblock Know me, respond to me: Benchmarking llms for dynamic user profiling and personalized responses at scale, 2025.
\newblock URL \url{https://arxiv.org/abs/2504.14225}.

\bibitem[Li et~al.(2025)Li, Xia, Yuan, Dong, Sha, Li, and Sui]{li-etal-2025-far}
Rui Li, Heming Xia, Xinfeng Yuan, Qingxiu Dong, Lei Sha, Wenjie Li, and Zhifang Sui.
\newblock How far are {LLM}s from being our digital twins? a benchmark for persona-based behavior chain simulation.
\newblock In Wanxiang Che, Joyce Nabende, Ekaterina Shutova, and Mohammad~Taher Pilehvar (eds.), \emph{Findings of the Association for Computational Linguistics: ACL 2025}, pp.\  15738--15763, Vienna, Austria, July 2025. Association for Computational Linguistics.
\newblock ISBN 979-8-89176-256-5.
\newblock \doi{10.18653/v1/2025.findings-acl.813}.
\newblock URL \url{https://aclanthology.org/2025.findings-acl.813/}.

\bibitem[{Meta}(2026)]{meta2026muse}
{Meta}.
\newblock Introducing {Muse}: The world's first personal {AI} agent built for everyone.
\newblock \url{https://about.fb.com/news/2026/09/introducing-muse-personal-ai-agent/}, September 2026.
\newblock Accessed: October 30, 2026.

\bibitem[{OpenAI}(2026)]{openai2026dots}
{OpenAI}.
\newblock Introducing dots.
\newblock \url{https://openai.com/index/introducing-dots/}, September 2026.
\newblock Accessed: October 30, 2026.

\bibitem[Park et~al.(2023)Park, O'Brien, Cai, Morris, Liang, and Bernstein]{park2023generative}
Joon~Sung Park, Joseph O'Brien, Carrie~Jun Cai, Meredith~Ringel Morris, Percy Liang, and Michael~S Bernstein.
\newblock Generative agents: Interactive simulacra of human behavior.
\newblock In \emph{Proceedings of the 36th annual acm symposium on user interface software and technology}, pp.\  1--22, 2023.

\bibitem[Park et~al.(2026)Park, Zou, Kamphorst, Egan, Shaw, Hill, Cai, Morris, Liang, Willer, and Bernstein]{park2026llmagentsgroundedselfreports}
Joon~Sung Park, Carolyn~Q. Zou, Jonne Kamphorst, Niles Egan, Aaron Shaw, Benjamin~Mako Hill, Carrie Cai, Meredith~Ringel Morris, Percy Liang, Robb Willer, and Michael~S. Bernstein.
\newblock Llm agents grounded in self-reports enable general-purpose simulation of individuals, 2026.
\newblock URL \url{https://arxiv.org/abs/2411.10109}.

\bibitem[Piao et~al.(2026)Piao, Yan, Zhang, Li, Yan, Lan, Lu, Zheng, Wang, Zhou, Gao, Xu, Zhang, Rong, Su, and Li]{piao2026agentsocietylargescalesimulationllmdriven}
Jinghua Piao, Yuwei Yan, Jun Zhang, Nian Li, Junbo Yan, Xiaochong Lan, Zhihong Lu, Zhiheng Zheng, Jing~Yi Wang, Di~Zhou, Chen Gao, Fengli Xu, Fang Zhang, Ke~Rong, Jun Su, and Yong Li.
\newblock Agentsociety: Large-scale simulation of llm-driven generative agents advances understanding of human behaviors and society, 2026.
\newblock URL \url{https://arxiv.org/abs/2502.08691}.

\bibitem[QI et~al.(2026)QI, Zhang, Zeng, Liu, Zhou, Miao, Yan, and Guan]{qi-etal-2026-beyond}
Yirui QI, Xiaoming Zhang, Ruilin Zeng, Mengyao Liu, Ziyi Zhou, Dezhuang Miao, Bingyu Yan, and Zhenyu Guan.
\newblock Beyond static persona consistency: Dynamic persona coherence in {LLM} role-playing.
\newblock In Maria Liakata, Viviane~P. Moreira, Jiajun Zhang, and David Jurgens (eds.), \emph{Proceedings of the 64th Annual Meeting of the {A}ssociation for {C}omputational {L}inguistics (Volume 1: Long Papers)}, pp.\  28942--28956, San Diego, California, United States, July 2026. Association for Computational Linguistics.
\newblock ISBN 979-8-89176-390-6.
\newblock \doi{10.18653/v1/2026.acl-long.1336}.
\newblock URL \url{https://aclanthology.org/2026.acl-long.1336/}.

\bibitem[Vezhnevets et~al.(2023)Vezhnevets, Agapiou, Aharon, Ziv, Matyas, Duéñez-Guzmán, Cunningham, Osindero, Karmon, and Leibo]{vezhnevets2023generativeagentbasedmodelingactions}
Alexander~Sasha Vezhnevets, John~P. Agapiou, Avia Aharon, Ron Ziv, Jayd Matyas, Edgar~A. Duéñez-Guzmán, William~A. Cunningham, Simon Osindero, Danny Karmon, and Joel~Z. Leibo.
\newblock Generative agent-based modeling with actions grounded in physical, social, or digital space using concordia, 2023.
\newblock URL \url{https://arxiv.org/abs/2312.03664}.

\bibitem[Vezhnevets et~al.(2025)Vezhnevets, Matyas, Cross, Paglieri, Chang, Cunningham, Osindero, Isaac, and Leibo]{vezhnevets2025multiactorgenerativeartificialintelligence}
Alexander~Sasha Vezhnevets, Jayd Matyas, Logan Cross, Davide Paglieri, Minsuk Chang, William~A. Cunningham, Simon Osindero, William~S. Isaac, and Joel~Z. Leibo.
\newblock Multi-actor generative artificial intelligence as a game engine, 2025.
\newblock URL \url{https://arxiv.org/abs/2507.08892}.

\bibitem[Wang et~al.(2025)Wang, Zhang, Yang, Chen, Tang, Zhang, Chen, Lin, Sun, Song, et~al.]{wang2025user}
Lei Wang, Jingsen Zhang, Hao Yang, Zhi-Yuan Chen, Jiakai Tang, Zeyu Zhang, Xu~Chen, Yankai Lin, Hao Sun, Ruihua Song, et~al.
\newblock User behavior simulation with large language model-based agents.
\newblock \emph{ACM Transactions on Information Systems}, 43\penalty0 (2):\penalty0 1--37, 2025.

\bibitem[Wu et~al.(2025)Wu, Wang, Yu, Zhang, Chang, and Yu]{wu2025longmemevalbenchmarkingchatassistants}
Di~Wu, Hongwei Wang, Wenhao Yu, Yuwei Zhang, Kai-Wei Chang, and Dong Yu.
\newblock Longmemeval: Benchmarking chat assistants on long-term interactive memory, 2025.
\newblock URL \url{https://arxiv.org/abs/2410.10813}.

\bibitem[Yuan et~al.(2024)Yuan, Sun, Li, Wang, Cao, and Li]{yuan2024personalizedlargelanguagemodel}
Ruifeng Yuan, Shichao Sun, Yongqi Li, Zili Wang, Ziqiang Cao, and Wenjie Li.
\newblock Personalized large language model assistant with evolving conditional memory, 2024.
\newblock URL \url{https://arxiv.org/abs/2312.17257}.

\bibitem[Zhou et~al.(2024)Zhou, Zhu, Mathur, Zhang, Yu, Qi, Morency, Bisk, Fried, Neubig, and Sap]{zhou2024sotopiainteractiveevaluationsocial}
Xuhui Zhou, Hao Zhu, Leena Mathur, Ruohong Zhang, Haofei Yu, Zhengyang Qi, Louis-Philippe Morency, Yonatan Bisk, Daniel Fried, Graham Neubig, and Maarten Sap.
\newblock Sotopia: Interactive evaluation for social intelligence in language agents, 2024.
\newblock URL \url{https://arxiv.org/abs/2310.11667}.

\bibitem[Zhou et~al.(2025)Zhou, Su, Feng, Zhou, Huang, Kao, Lynch, Volkova, Wu, Woolley, Zhu, and Sap]{zhou-etal-2025-sotopia}
Xuhui Zhou, Zhe Su, Sophie Feng, Jiaxu Zhou, Jen-tse Huang, Hsien-Te Kao, Spencer Lynch, Svitlana Volkova, Tongshuang Wu, Anita Woolley, Hao Zhu, and Maarten Sap.
\newblock {SOTOPIA}-s4: a user-friendly system for flexible, customizable, and large-scale social simulation.
\newblock In Nouha Dziri, Sean~(Xiang) Ren, and Shizhe Diao (eds.), \emph{Proceedings of the 2025 Conference of the Nations of the Americas Chapter of the Association for Computational Linguistics: Human Language Technologies (System Demonstrations)}, pp.\  350--360, Albuquerque, New Mexico, April 2025. Association for Computational Linguistics.
\newblock ISBN 979-8-89176-191-9.
\newblock \doi{10.18653/v1/2025.naacl-demo.30}.
\newblock URL \url{https://aclanthology.org/2025.naacl-demo.30/}.

\bibitem[Zhou et~al.(2026)Zhou, Sun, Du, Liu, Sun, Ma, Wu, Yang, and Sap]{zhou2026odyssim}
Xuhui Zhou, Weiwei Sun, Weihua Du, Jiarui Liu, Haojia Sun, Qianou Ma, Tongshuang Wu, Yiming Yang, and Maarten Sap.
\newblock Odyssim: Building foundation models for human behavior simulation.
\newblock \emph{arXiv preprint arXiv:2606.14199}, 2026.

\end{thebibliography}
\bibliographystyle{iclr2027_conference}

\appendix
\section{Longitudinal Evaluation: Implementation Details}
\label{app:eval_impl}
 
This appendix documents the reference implementation of the evaluation framework in Section~\ref{sec:eval}.
 
\subsection{Pipeline overview}
The pipeline operates on a parsed simulation trajectory (per-actor, per-time-slot records) and proceeds in four stages.
 
\textbf{Event-thread extraction.} Atomic events are extracted from the raw trajectory, then linked into candidate threads using shared actors, shared objects or goals, explicit references, causal compatibility, and temporal plausibility. Linking allows an event to be assigned to a \texttt{NEW\_THREAD} or left as \texttt{UNCERTAIN} rather than forcing every event into an existing thread. Each resulting thread is captured as an event card recording its actors, type, initiation time, core object, state sequence, current status, open dependencies, and supporting evidence. EEE and ECD are both computed from these cards.
 
\textbf{Judge interface.} Scoring is abstracted behind a common \texttt{Judge} interface, and every run reported in this paper uses \texttt{LLMJudge} rather than the deterministic \texttt{HeuristicJudge} kept for offline unit tests. This means every \emph{semantic label} -- Pass A/B transition labels, ECD episode dimensions, EEE dimensions, CSRC dimensions, and the per-item NDD decisions -- is produced by the judge model, with response caching, retry handling, and batched/threaded requests for throughput. It does not mean every \emph{number} in the report is itself a model output: taxonomy-mode Transition Alignment Scores are a fixed deterministic lookup applied to the Pass A/B labels (Appendix~\ref{app:eval_idc_def}), and all rates, counts, and descriptive statistics are computed from judged labels in Python rather than requested from the model directly.
 
\textbf{Per-block scoring.} IDC scoring follows the three-pass protocol of Appendix~\ref{app:eval_idc_def}. ECD scoring computes the five episode dimensions and the Perturbation Win Rate against six corruption operators (unrelated response, dependency violation, commitment amnesia, premature closure, mechanical overreaction, mechanical inertia), each applied at the episode level. EEE scoring computes Event Initiation Plausibility, Event Progression Coherence, and Event Status Appropriateness per Event Card. A dedicated LLM diagnostics stage judges the Context-Sensitive Response Calibration dimensions and the per-item decisions underlying all five NDD diagnostics (Appendix~\ref{app:eval_ndd_def}) using earlier judge outputs and retrieved downstream evidence. Deterministic lexical versions of both exist in the code as legacy scaffolding for the offline \texttt{HeuristicJudge} path, but every reported CSRC and NDD number aggregates the LLM-judged version.
 
\textbf{Abstention and reporting.} Per-item judgments the pipeline cannot support with sufficient evidence (e.g.\ \textsc{insufficient\_evidence} Pass A/B labels, undecidable NDD windows) are quarantined rather than merged or imputed. Available eligibility denominators are reported in Table~\ref{tab:eval_coverage}. Unavailable counts are marked NR. Scores remain conditional on eligibility, and differing abstention patterns can affect comparisons. One exception: IDC Macro and the three pooled IDC rates -- Unsupported Drift Rate, Missed Adaptation Rate, and Spillover Rate -- return 0.0 rather than N/A by convention when the relevant denominator is empty. The denominator counts eligible actor--facet transitions for IDC Macro, eligible \textsc{stay} cases for Unsupported Drift Rate, eligible \textsc{should\_change} cases for Missed Adaptation Rate, and triggered transitions for Spillover Rate. The denominator is reported alongside each value, so a 0.0 next to a denominator of zero should be read as ``not evaluated'' rather than ``no failures observed'' (for the rates) or ``perfectly consistent'' (for IDC Macro). A final aggregation step assembles all blocks into a single report, which we render as the tables in this appendix.
 
\subsection{Formal metric definitions}
\label{app:eval_idc_def}

\paragraph{Individual Dynamic Consistency (IDC).}
An actor's situation can change without every aspect of its behavior changing with it. IDC assesses whether transitions in communication style, behavioral patterns, goals and commitments, and interests and preferences are supported by intervening events. It also checks whether changes spread to unrelated facets or whether the actor fails to adapt when the accumulated context provides strong reason to do so. Each eligible actor--facet transition receives a Transition Alignment Score (TAS) from $0$ to $4$. We report the average across actors as
\begin{equation}
\mathrm{IDC\ Macro}
=
\frac{100}{|\mathcal A|}
\sum_{i\in\mathcal A}
\frac{1}{|\mathcal U_i|}
\sum_{u\in\mathcal U_i}
\frac{s_u}{4},
\end{equation}
where $i$ denotes an actor, $\mathcal A$ is the set of actors with at least one eligible transition, $\mathcal U_i$ is the set of eligible actor--facet transitions for actor $i$, $u$ denotes one such transition, and $s_u\in\{0,1,2,3,4\}$ is its alignment score. The vertical bars give the number of elements in each set, and the factors $1/4$ and $100$ convert the score to a $0$--$100$ scale. A higher IDC Macro means that an actor's observed stability and changes are more consistently supported by its history.

To obtain $s_u$, an expected-update pass sees the pre-event state and an approximate trigger clause extracted from the current action, and labels the facet \textsc{stay}, \textsc{may\_change}, \textsc{should\_change}, or \textsc{insufficient\_evidence}. A separate observed-update pass sees the pre/post trajectory and labels the observed change \textsc{none}, \textsc{minor}, \textsc{material}, or \textsc{insufficient\_evidence}, with supporting evidence. Both passes query the judge. A deterministic lookup converts the two labels into TAS for \textsc{stay} and \textsc{should\_change}, and for \textsc{may\_change} when the taxonomy mode is used. The alternative mode queries the judge for the plausibility of a realized \textsc{may\_change} transition. The trigger clause is split from action text using predefined connective words. If no split is found, the full action is used for both trigger and response, so the expected-update pass is not guaranteed to exclude response information.

We also report Unsupported Drift Rate (UDR) for material changes without sufficient support, Missed Adaptation Rate (MAR) for failures to update when the context strongly supports adaptation, and Spillover Rate for unjustified changes in unrelated facets:
\begin{gather*}
\mathrm{UDR} = \frac{\#\text{unsupported material changes}}{\#\text{eligible \textsc{stay} cases}},\\[4pt]
\mathrm{MAR} = \frac{\#\{\text{\textsc{should\_change} cases with observed change \textsc{none} or \textsc{minor}}\}}{\#\text{eligible \textsc{should\_change} cases}}.
\end{gather*}
Here $\#$ denotes a count. Spillover Rate divides the number of triggered transitions with an unjustified change in an unrelated facet by the number of triggered transitions eligible for that judgment. Lower values are better for all three rates. UDR, MAR, and Spillover Rate pool eligible cases across actors, whereas IDC Macro first averages within actors. MAR measures the magnitude of an observed update, not whether a material update moved in the correct direction. The taxonomy TAS for \textsc{should\_change} has the same limitation.

\label{app:eval_pass_c}
Because many triggers admit multiple reasonable continuations, the boundary between \textsc{may\_change} and \textsc{should\_change} can be unreliable. Once per evaluation, the pipeline runs the expected-update pass on a sampled batch of 24 transitions by default and measures the share of decidable facet labels that are \textsc{may\_change}. If this share is at least 50\%, \textsc{may\_change} facets use the realized-transition-plausibility judge rather than the fixed taxonomy lookup for the rest of the run. Otherwise, the taxonomy lookup is retained. The \textsc{stay} and \textsc{should\_change} labels always use the deterministic lookup. The report records the mode selected for each run.

\paragraph{Emergent Event Evaluation (EEE).}
The simulator does not follow a fixed list of events to complete, so we identify event threads from actions concerning a continuing goal, problem, or shared activity. EEE evaluates three questions about each thread. Event Initiation Plausibility (EIP) asks whether its beginning is grounded in the actor's state or the world. Event Progression Coherence (EPC) asks whether later developments respect prior outcomes, dependencies, and elapsed time. Event Status Appropriateness (ESA) asks whether its current status, including remaining open or waiting, follows from its history. This last measure avoids treating rapid completion as inherently better than a justified ongoing process. Each measure averages its applicable rubric scores and rescales them to $0$--$100$:
\begin{equation}
Q_m(e)
=
\frac{100}{4|D_m(e)|}
\sum_{d\in D_m(e)} s_{e,d},
\qquad
m\in\{\mathrm{EIP},\mathrm{EPC},\mathrm{ESA}\},
\end{equation}
where $e$ denotes an event thread, $m$ identifies one of the three EEE measures, and $Q_m(e)$ is that measure's score for thread $e$. The set $D_m(e)$ contains the applicable rubric criteria, $d$ denotes one criterion, and $s_{e,d}\in\{0,1,2,3,4\}$ is its score. The vertical bars give the number of applicable criteria. Higher EIP, EPC, and ESA scores indicate stronger support for the event's initiation, progression, and current status, respectively. EIP assesses contextual grounding, actor compatibility, consistency with established history, and potential future consequences. EPC assesses causal coherence, state continuity, timing, and dependency handling. ESA assesses whether the status and, for terminated events, the resolution fit the thread's history. We average eligible thread scores for each measure.

\paragraph{Event Collaboration Dynamics (ECD).}
An event thread can progress coherently overall while one participant forgets a commitment or acts before another has completed a required step. ECD therefore evaluates decision episodes within multi-actor threads. For each episode, it considers the current event state, the focal actor's role and commitments, the relevance of its response to the trigger, dependencies among participants, and whether the response remains consistent with subsequent steps. Acting and waiting can both be appropriate when supported by the actor's information and the event history. We summarize these five dimensions with Episode Collaboration Plausibility (ECP):
\begin{equation}
\mathrm{ECP}(c)
=
\frac{100}{4|K_c|}
\sum_{k\in K_c} r_{c,k},
\end{equation}
where $c$ denotes a collaboration episode, $K_c$ is the set of dimensions with sufficient evidence, $k$ denotes one such dimension, and $r_{c,k}\in\{0,1,2,3,4\}$ is its score. The vertical bars give the number of scored dimensions. A higher ECP score means the participant's response better respects the shared event state, commitments, dependencies, and subsequent trajectory. We average episode scores within each event thread before averaging across threads, so a long event does not dominate the result. Context-Sensitive Response Calibration (CSRC) additionally checks whether responses differ appropriately across episodes with different triggers, accumulated evidence, commitments, dependencies, and actor histories. A higher CSRC score indicates more appropriate context sensitivity. ECD measures coordination within shared events, not general interpersonal relationship quality.

\paragraph{Non-degeneracy Diagnostics (NDD).}
\label{app:eval_ndd_def}
Strong scores on eligible transitions and events do not ensure that a full trajectory develops meaningfully. A simulator might avoid inconsistencies by producing almost no consequential activity, repeat similar interactions without lasting effects, or continually introduce weakly grounded events. NDD therefore reports five complementary diagnostics. Meaningful Event-Thread Rate counts judge-confirmed meaningful threads per $1{,}000$ observed actor-slots, where one actor-slot is one actor in one simulation time slot. A low rate can indicate stagnation, but a high rate alone is not evidence of quality. Supported Novelty Ratio is the share of newly initiated threads whose grounding and actor-compatibility scores are both at least $3$ out of $4$. Repetition/Near-Duplicate Rate is the share of decision windows that repeat an earlier pattern without a new consequence. Consequence Rate is the share of meaningful events with a downstream effect found in later observations. History-Dependence Rate is the share of decisions with relevant prior history that use it appropriately. We compute the five diagnostics as follows, using MTR for Meaningful Event-Thread Rate, SNR for Supported Novelty Ratio, RNR for Repetition/Near-Duplicate Rate, CR for Consequence Rate, and HDR for History-Dependence Rate:
\begin{equation}
\begin{aligned}
\mathrm{MTR} &= 1000\,\frac{\#\text{meaningful threads}}{\#\text{observed actor-slots}},\\[3pt]
\mathrm{SNR} &= 100\,\frac{\#\text{supported new threads}}{\#\text{eligible new threads}},\\[3pt]
\mathrm{RNR} &= 100\,\frac{\#\text{repeated windows}}{\#\text{decidable windows}},\\[3pt]
\mathrm{CR} &= 100\,\frac{\#\text{threads with downstream effects}}{\#\text{consequence-eligible threads}},\\[3pt]
\mathrm{HDR} &= 100\,\frac{\#\text{decisions appropriately using history}}{\#\text{history-eligible decisions}},
\end{aligned}
\end{equation}
where $\#$ denotes a count. An eligible new thread is judged to have begun within the observed window and has both grounding and actor-compatibility scores. It is supported when both scores are at least $3$ out of $4$. The repetition denominator contains windows for which repetition can be judged. Consequence-eligible threads are meaningful threads for which a later effect can be assessed. History-eligible decisions have relevant prior history and a decidable judgment of whether the decision uses it appropriately. MTR is a density per $1{,}000$ actor-slots, while the other four measures are percentages. Higher SNR, CR, and HDR and lower RNR are preferable. MTR has no universal target. We interpret the measures together and do not assume a universal target frequency for events or state changes. Undecidable cases are excluded from the relevant denominator, and we report eligibility alongside each rate in Appendix~\ref{app:eval_coverage}.

\paragraph{Evaluation reliability.}
Post-hoc judgments risk accepting an observed response merely because a plausible explanation can be constructed for it. We require the judge to ground its scores in trajectory evidence and compare original collaboration episodes with controlled corruptions, including forgotten commitments and violated dependencies. Perturbation Win Rate (PWR) records how often the judge scores the original episode higher. Following the implemented aggregation, we average these comparisons within threads, then across threads for each corruption type, and finally across corruption types:
\begin{equation}
\begin{aligned}
\mathrm{PWR}
&= \frac{1}{|\mathcal C|}\sum_{c\in\mathcal C}
\frac{1}{|\mathcal T_c|}\sum_{t\in\mathcal T_c}\\
&\qquad\times\frac{1}{|\mathcal P_{c,t}|}\sum_{j\in\mathcal P_{c,t}}
\mathbf{1}\!\left[q(x_j)>q(\tilde{x}_j)\right],
\end{aligned}
\end{equation}
where $\mathcal C$ is the set of eligible corruption types, $\mathcal T_c$ is the set of eligible event threads for type $c$, and $\mathcal P_{c,t}$ is the set of eligible original--corrupted episode pairs in thread $t$ for type $c$. The index $j$ identifies a pair, $x_j$ is its original episode, $\tilde{x}_j$ is its corrupted version, and $q$ is the judge's score on dimensions available for both. The indicator $\mathbf{1}[\cdot]$ is $1$ when the original scores higher and $0$ otherwise, so ties are not wins. PWR ranges from $0$ to $1$. A higher value indicates that the judge better distinguishes coherent episodes from controlled errors. It checks the evaluator, not the quality of a generated trajectory.

\subsection{Aggregation and coverage across conditions}
\label{app:eval_coverage}

Each condition in Table~\ref{tab:baseline_comparison} contributes one
trajectory. No metric is averaged across independent simulation
runs. The within-trajectory aggregation procedures are:

\begin{itemize}
    \item \textbf{IDC Macro:} average eligible transition--facet
    TAS values within each actor, then average equally across
    eligible actors and scale to 0--100. Unsupported Drift Rate,
    Missed Adaptation Rate, and Spillover Rate instead pool their respective
    eligible cases across actors.

    \item \textbf{Episode Collaboration Plausibility:} average
    applicable dimensions within each decision episode, average
    episode scores within each thread, then average equally
    across scorable threads and scale to 0--100.

    \item \textbf{Context-Sensitive Response Calibration:}
    average applicable calibration dimensions within each actor,
    then average equally across eligible actors and scale to
    0--100.

    \item \textbf{Perturbation Win Rate:} compare original and
    corrupted episodes using dimensions scored in both versions.
    Average binary wins over eligible episodes within each thread,
    then equally across eligible threads for each corruption
    type, and finally equally across applicable corruption types.
    Ties are not wins.

    \item \textbf{EEE:} for each of EIP, EPC, and ESA, average
    applicable dimensions within each Event Card, then average
    equally across cards eligible for that metric and scale to
    0--100.

    \item \textbf{NDD:} Meaningful Event Threads divides the number
    of confirmed meaningful threads by observed actor-slots
    and multiplies by 1,000. Supported Novelty and Consequence
    pool eligible threads. Near-Duplicate Actions pools decidable decision
    windows. History-Dependence pools decisions with relevant
    history and a decidable history-use judgment. These four
    proportions are expressed on 0--100 scales.
\end{itemize}

Table~\ref{tab:eval_coverage} reports coverage separately from
scores. Counts of actors, threads, episodes, and decision windows
are within-trajectory evaluation units, not independent simulation
replicates. Eligibility differs across metrics, so high scores
should be interpreted alongside their coverage.

\begin{table}[t]
\centering
\footnotesize
\setlength{\tabcolsep}{4pt}
\caption{Coverage and available eligibility denominators for
the five longitudinal evaluations. NR means not reported in
the available Markdown export. It does not mean zero or
ineligible.}
\label{tab:eval_coverage}
\begin{tabular}{lrrrrr}
\toprule
Evaluation unit or denominator & \framework{} & Direct & Guided & Per-Actor & Summary \\
\midrule
Observed actor-slots & 3600 & 3600 & 3600 & 3600 & 3600 \\
Accepted threads & 84 & 25 & 45 & 36 & 28 \\
\midrule
IDC eligible \textsc{stay} cases & 5696 & 4223 & NR & NR & 7185 \\
IDC eligible \textsc{should\_change} cases & 3373 & 5467 & NR & NR & 3249 \\
IDC triggered transitions & 1909 & 2940 & NR & NR & 1922 \\
\midrule
ECD scorable threads & 26 & 5 & 13 & 14 & 9 \\
ECD decision episodes & 1550 & 339 & 871 & 1898 & 1678 \\
CSRC eligible actors & 10 & 10 & 10 & 10 & 10 \\
\midrule
EIP eligible cards & 79 & 24 & 45 & 36 & 28 \\
EPC eligible cards & 84 & 25 & 45 & 36 & 27 \\
ESA eligible cards & 51 & 7 & 16 & 18 & 13 \\
\midrule
Meaningfulness-decidable threads & 77 & 14 & 37 & 31 & 25 \\
Confirmed meaningful threads & 77 & 14 & 37 & 30 & 23 \\
Novelty-eligible threads & 50 & 4 & 13 & 14 & 10 \\
Consequence-eligible threads & 68 & 14 & 37 & 30 & 23 \\
Repetition-decidable windows & 2659 & 3127 & 3423 & 3148 & 3041 \\
History-use-decidable decisions & 2621 & 3105 & 3400 & 3133 & 2995 \\
\bottomrule
\end{tabular}
\end{table}

The ECD episode counts are not universal denominators for
Perturbation Win Rate: individual comparisons require jointly
scored dimensions, and mechanical-inertia corruptions additionally
require prior context. Perturbation results therefore follow
the hierarchical aggregation described above.

\subsection{Detailed aggregate report for Direct}
\label{app:eval_example}
Table~\ref{tab:eval_example} shows the detailed aggregate output of the pipeline underlying the \textbf{Direct} column of Table~\ref{tab:baseline_comparison} (evaluation protocol v5, LLM judge). We include it here to illustrate the reporting format, denominators, and calibration diagnostics the pipeline tracks. The summary values correspond to the Direct column of Table~\ref{tab:baseline_comparison}, subject to rounding. This is not an independent exploratory configuration.
 
\begin{table}[h]
\centering
\footnotesize
\caption{Detailed aggregate evaluation report for the \textbf{Direct} baseline (evaluation protocol v5, LLM judge). Thread extraction: \texttt{semantic-subevents-v3} (LLM judge), 25 accepted threads (largest: 239 actions). Uncertainty policy: \texttt{quarantine-v3-abstention-validation}. Among event actions, 2,298 were accepted and 2,408 were excluded as uncertain (excluded from ECD/EEE). Accepted subevents: 5489/13621. Source-level accepted/routine/uncertain counts may overlap.}
\label{tab:eval_example}
\begin{tabular}{llrl}
\toprule
Block & Metric & Value & Range \\
\midrule
IDC & IDC Macro & 56.51 & 0--100 \\
IDC & Unsupported Drift Rate & 0.615 & 0--1 \\
IDC & Missed Adaptation Rate & 0.334 & 0--1 \\
IDC & Spillover Rate & 0.475 & 0--1 \\
IDC & Pass-C mode & taxonomy & -- \\
\midrule
ECD & Collaboration Plausibility & 95.92 & 0--100 \\
ECD & D1 / D2 / D3 / D4 / D5 & 4.00 / 3.98 / 3.99 / 3.62 / 3.60 & 0--4 each \\
ECD & Context-Sensitive Resp.\ Calibration & 84.0 & 0--100 \\
ECD & Perturbation Win Rate & 0.896 & 0--1 \\
\midrule
EEE & Event Initiation Plausibility & 77.34 & 0--100 \\
EEE & Event Progression Coherence & 71.33 & 0--100 \\
EEE & Event Status Appropriateness & 82.14 & 0--100 \\
\midrule
NDD & Meaningful Event Threads & 3.889 & per 1{,}000 actor-slots \\
NDD & Supported Novelty Ratio & 100.0 & 0--100 \\
NDD & Near-Duplicate Actions & 73.52 & \% \\
NDD & Consequence Rate & 100.0 & 0--100 \\
NDD & History-Dependence Rate & 78.68 & 0--100 \\
\midrule
Calib. & Human agreement & N/A & -- \\
Calib. & Taxonomy vs.\ plausibility closeness & 77.5 & \% \\
\bottomrule
\end{tabular}
\end{table}
 
For reference, this Direct run had 5 collaboration-eligible threads out of 25 (individual, single-actor event coherence on the remaining 20 threads was 82.5/100 under a separate rubric). It had 339 ECD decision episodes over those 5 threads (dimension coverage 339/339/339/339/337 for D1--D5). The EEE card counts were 24, 25, and 7 for EIP, EPC, and ESA, respectively. CSRC covered 10/10 eligible actors. NDD denominators were 3600 observed actor-slots, 14 meaningfulness-decidable threads (14 meaningful), 4 novelty-eligible threads (11 observed new threads), 14 consequence-eligible threads, and 3105/3127 decisions decidable for history-use and repetition respectively (56 history-citation and 56 history-row abstentions, 1089 downstream-retrieval batches with 8 abstentions, and 11 event-diagnostic abstentions). A purely descriptive statistic, exact normalized action repetition, was separately computed at 65.4\% and is reported for context rather than as a scored metric. All LLM-judged scores in this table remain model-based semantic audits rather than human validation.

\section{Additional SOTOPIA Analyses}
\label{app:sotopia_analyses}

We conduct two analyses involving SOTOPIA \citep{zhou2024sotopiainteractiveevaluationsocial}. The first scores SOTOPIA episodes with our longitudinal metrics. The second scores our 60-day trajectories with an adapted version of SOTOPIA's native evaluator. Because the data and scoring procedures differ, the resulting scores are not directly comparable.

\subsection{Longitudinal Metrics on SOTOPIA Episodes}
\label{app:sotopia_sample}

We draw a fixed-seed ($\mathrm{seed}=42$) stratified random sample of 160 episodes from 7{,}049 usable SOTOPIA episodes, allocating 10 episodes to each of 16 ordered acting-model pairs. The pipeline completed 159/160 episodes (one failure from a thread-linking error, retained rather than imputed), spanning 76 distinct environments and 1{,}312 retained actions.

A zero-denominator convention in the underlying pipeline writes 0 for IDC when no actor--facet is eligible, and for the three drift/adaptation/spillover rates when no eligible case exists (Appendix~\ref{app:eval_impl}). Excluding these 27--33 zero-denominator episodes per metric, rather than treating the default as a true zero, raises IDC Macro from 48.21 to 58.08, Unsupported Drift Rate from 0.181 to 0.221, Missed Adaptation Rate from 0.617 to 0.778, and Spillover Rate from 0.130 to 0.164. 
Table~\ref{tab:sotopia_main} reports only the eligibility-adjusted values.

Table~\ref{tab:sotopia_pairs} breaks the block-level scores down by ordered actor-model pair. We caution against over-interpreting this breakdown: episode counts per pair are small (typically $n \le 10$ eligible episodes) and episodes sharing an environment or character configuration may be dependent, so the reported SDs describe between-episode dispersion rather than a basis for significance testing. Original SOTOPIA profiles, goals, and human ratings were not supplied to the metric judge, and no cross-metric overall score is computed given the differing scales and eligibility across metrics.
\begin{table}[t]
\centering
\footnotesize
\caption{Longitudinal metrics applied to a stratified sample of 160 SOTOPIA episodes (10 per ordered acting-model pair, 16 pairs), of which 159 completed the pipeline (Table~\ref{tab:sotopia_pairs} reports per-pair completion, with one pair completing 9/10). Higher is better except where marked $\downarrow$. Statistics are equal-episode-weighted means and sample SDs over eligible nonmissing scores from the 159 completed episodes. Valid $n$ varies across metrics because zero-denominator and insufficient-evidence cases are excluded rather than assigned a value (Appendix~\ref{app:eval_impl}).}
\label{tab:sotopia_main}
\begin{tabular}{lcccc}
\toprule
Metric & Scale & Valid $n$ & Mean & SD \\
\midrule
IDC Macro & 0--100 & 132 & 58.08 & 15.06 \\
Unsupported Drift Rate $\downarrow$ & 0--1 & 130 & 0.221 & 0.186 \\
Missed Adaptation Rate $\downarrow$ & 0--1 & 126 & 0.778 & 0.205 \\
Spillover Rate $\downarrow$ & 0--1 & 126 & 0.164 & 0.199 \\
\midrule
Episode Collab.\ Plausibility & 0--100 & 87 & 88.48 & 7.95 \\
Context-Sensitive Resp.\ Calib. & 0--100 & 128 & 69.56 & 15.02 \\
Perturbation Win Rate & 0--1 & 87 & 0.950 & 0.050 \\
\midrule
Event Initiation Plausibility & 0--100 & 122 & 68.70 & 17.46 \\
Event Progression Coherence & 0--100 & 119 & 84.26 & 13.03 \\
Event Status Appropriateness & 0--100 & 102 & 95.42 & 13.10 \\
\midrule
Meaningful Thread Rate & per 1{,}000 actor-slots & 110 & 230.32 & 237.83 \\
Supported Novelty Ratio & 0--100 & 40 & 66.67 & 45.76 \\
Repetition Rate $\downarrow$ & 0--100 & 138 & 27.24 & 26.39 \\
Consequence Rate & 0--100 & 82 & 100.00 & 0.00 \\
History-Dependence Rate & 0--100 & 128 & 69.41 & 30.10 \\
\bottomrule
\end{tabular}
\end{table}

\begin{table}[t]
\centering
\footnotesize
\setlength{\tabcolsep}{3pt}
\caption{Block-level scores by ordered acting-model pair (Actor~1, Actor~2), mean $\pm$ sample SD ($n$), 0--100 scales. Because SOTOPIA interactions are dyadic and asymmetric, pair-level scores reflect the joint episode and should not be attributed to either actor model alone.}
\label{tab:sotopia_pairs}
\resizebox{\linewidth}{!}{\begin{tabular}{llrlllll}
\toprule
Actor 1 & Actor 2 & Compl.\ & IDC & ECD & EIP & EPC & ESA \\
\midrule
gpt-3.5-turbo & gpt-3.5-turbo & 10/10 & 56.3±15.2(10) & 89.2±6.9(9) & 79.2±13.0(9) & 92.4±7.7(9) & 100.0±0.0(6) \\
gpt-3.5-turbo & gpt-4 & 10/10 & 60.1±12.0(10) & 91.9±4.3(9) & 69.6±16.7(10) & 95.0±4.7(10) & 94.4±9.1(9) \\
gpt-3.5-turbo & llama-2-70b-chat & 10/10 & 57.8±11.9(9) & 87.9±7.1(9) & 70.1±23.2(9) & 86.5±9.6(9) & 92.9±12.2(7) \\
gpt-3.5-turbo & mpt-30b-chat & 10/10 & 71.4±7.2(4) & N/A(0) & 59.4±18.8(4) & 79.2±26.0(3) & 96.9±6.3(4) \\
gpt-4 & gpt-3.5-turbo & 10/10 & 53.8±15.1(10) & 92.9±5.1(7) & 67.4±22.2(8) & 89.1±12.5(8) & 100.0±0.0(6) \\
gpt-4 & gpt-4 & 10/10 & 56.8±8.3(10) & 93.5±7.0(9) & 74.2±16.5(9) & 92.3±6.8(10) & 98.4±2.9(8) \\
gpt-4 & llama-2-70b-chat & 10/10 & 56.4±14.2(7) & 88.5±7.6(4) & 60.9±25.6(8) & 73.0±11.9(8) & 100.0±0.0(6) \\
gpt-4 & mpt-30b-chat & 10/10 & 62.5±29.3(4) & 90.9±4.4(3) & 64.6±21.5(6) & 89.9±9.1(6) & 87.5±30.6(6) \\
llama-2-70b-chat & gpt-3.5-turbo & 10/10 & 57.2±13.0(10) & 86.7±8.2(8) & 67.3±16.3(9) & 83.8±12.6(9) & 95.8±9.3(7) \\
llama-2-70b-chat & gpt-4 & 10/10 & 58.3±11.6(9) & 84.9±7.6(8) & 71.8±12.0(10) & 81.4±7.7(10) & 90.0±24.2(10) \\
llama-2-70b-chat & llama-2-70b-chat & 10/10 & 60.9±10.5(9) & 76.7±15.0(5) & 66.1±8.9(6) & 76.0±16.0(6) & 93.8±10.5(6) \\
llama-2-70b-chat & mpt-30b-chat & 10/10 & 54.5±37.2(5) & 90.7±1.6(2) & 65.3±16.8(3) & 83.3±14.4(3) & 91.7±14.4(3) \\
mpt-30b-chat & gpt-3.5-turbo & 10/10 & 53.8±12.2(10) & 87.3±1.0(3) & 69.8±19.6(9) & 77.7±14.7(8) & 100.0±0.0(6) \\
mpt-30b-chat & gpt-4 & 10/10 & 63.1±18.1(10) & 91.5±3.4(2) & 72.3±20.7(7) & 82.1±15.6(6) & 92.9±18.9(7) \\
mpt-30b-chat & llama-2-70b-chat & 9/10 & 54.5±15.3(9) & 88.4±4.6(4) & 65.4±14.6(8) & 83.0±10.7(7) & 95.0±11.2(5) \\
mpt-30b-chat & mpt-30b-chat & 10/10 & 61.5±17.5(6) & 84.4±10.7(5) & 63.8±11.9(7) & 71.6±16.0(7) & 100.0±0.0(6) \\
\bottomrule
\end{tabular}}
\end{table}

\subsection{Native SOTOPIA Evaluation of Our Trajectories}
\label{app:sotopia_adapter}

For this analysis, we adapt SOTOPIA's native \texttt{EpisodeLLMEvaluator} \citep{zhou2024sotopiainteractiveevaluationsocial} (repository commit \texttt{a0aaafb440e570e5e61b7c44a44e5e417c545383}).

We treat each complete 60-day trajectory as one multi-actor episode and map its ten actors and narrated actions to the native evaluator's agent and message representations. We retain the seven upstream \texttt{SotopiaDimensions} rubric definitions and native generation and parsing pipeline, while extending the response schema to record nullable scores, evidence status, source-action citations, and limitations. Because our trajectories do not include original SOTOPIA character profiles, assigned goals, or confidentiality specifications, Believability is marked as based on partial evidence, while Goal and Secret scores are set to N/A and excluded from aggregation. A compatibility patch provides explicit ten-actor key mapping, current score-object extraction, and failure propagation. Consequently, these results use an adapted native evaluator and are not directly comparable to scores from the unmodified SOTOPIA benchmark.

\begin{table}[t]
\centering
\footnotesize
\caption{Adapted native SOTOPIA evaluator scores (actor mean $\pm$ sample SD, $n=10$ actors) on five 60-day, 10-actor, 3{,}600-action trajectories, each scored in one joint GPT-4.1 evaluator call. Believability is marked with an asterisk because original character profiles were unavailable, so scores reflect partial evidence only. Secret and Goal are N/A throughout because our setting does not define assigned secrets or goal targets. Because each condition has a single joint evaluation, actor-level SD is descriptive (dispersion across actors within one run) and does not indicate run-to-run reliability or statistical significance.}
\label{tab:sotopia4_native}
\resizebox{\linewidth}{!}{\begin{tabular}{lccccc}
\toprule
Metric & \framework{} & Direct & Guided & Per-Actor & Summary \\
\midrule
Believability (0--10) & 9.00$\pm$0.00* & 9.00$\pm$0.00* & 9.00$\pm$0.00* & 9.00$\pm$0.00* & 9.00$\pm$0.00* \\
Relationship ($-$5--5) & 0.00$\pm$0.00 & 1.00$\pm$0.00 & 0.00$\pm$0.00 & 1.00$\pm$0.00 & 1.40$\pm$0.52 \\
Knowledge (0--10) & \textbf{7.50}$\pm$0.71 & 6.60$\pm$0.97 & 6.80$\pm$0.63 & 7.40$\pm$0.52 & 6.80$\pm$0.63 \\
Secret ($-$10--0) & N/A & N/A & N/A & N/A & N/A \\
Social rules ($-$10--0) & 0.00$\pm$0.00 & 0.00$\pm$0.00 & 0.00$\pm$0.00 & 0.00$\pm$0.00 & 0.00$\pm$0.00 \\
Financial/material ($-$5--5) & N/A & 0.00$\pm$0.00 & N/A & N/A & N/A \\
Goal (0--10) & N/A & N/A & N/A & N/A & N/A \\
\bottomrule
\end{tabular}}
\end{table}

\section{System Optimization}
\label{app:system_optimization}
The entire simulation framework is implemented using LangGraph.\footnote{\url{https://github.com/langchain-ai/langgraph}} We optimize model calls, prompt construction, parallel execution, and state management to reduce cost and generation time.

\paragraph{Model calls and inputs.}
We reduce estimated API cost and generation time by making fewer and smaller model calls. Fewer calls reduce token usage and sequential waiting, while smaller inputs avoid repeatedly processing information unrelated to the task. In our framework, we remove calls whose outputs are unused, run daily life planning only in the final slot of each simulated day, and call the model for world updates only when the judge identifies realized action effects. Routing, identifier assignment, and trajectory recording are handled by Python. Each component receives task-specific views of the relevant actor, goal, activity, and world information.

\paragraph{History compaction and caching.}
We reduce estimated input-token cost by limiting prompt growth and increasing cache reuse. Compacting superseded progress records reduces repeated input, while keeping reusable content stable across calls improves caching. We apply this to Stage Goal histories by retaining the latest complete progress update for each goal within the older portion of the prompt history. Recent records remain unchanged, and the full original event stream is preserved for auditing. We also separate stable context, older history, and current-slot updates. For components checked through replay, stable content is placed in \texttt{instructions}, with cache routing distinguished by actor, component, and prompt version. By default, the cutoff between older and recent history advances once per simulated day.

\paragraph{Parallel actor execution.}
We reduce elapsed generation time by running actor branches concurrently. Within a slot, actors use their own states and observations from the same frozen environment snapshot, so they do not need to wait for one another's current-slot outputs. Their model calls can therefore overlap. The ten-actor configuration uses ten workers by default. Updates are merged after all actors reach a synchronization barrier, and preassigned identifiers determine the order of trajectory records. This optimization changes execution scheduling without removing model calls.

\paragraph{State replay and audit logging.}
We reduce CPU and memory overhead in state replay, checkpointing, and recovery. Indexed replay avoids processing the full history for every event, while separate audit storage avoids repeatedly copying and encoding accumulated call records. For goal-event replay, we maintain indexes over event identifiers, actor sequences, and goal histories and update the local replay state in place. This removes repeated full-history validation and registry copying. Actor branches do not copy accumulated LLM call records. Historical call text is serialized once and appended to a separate audit file. Full checkpoints are still written, using previously serialized data and streamed output to reduce repeated encoding and intermediate memory use. During recovery, call records are read sequentially, with only the required usage information retained in memory.

\paragraph{Validation scope.}
Prompt compaction and changes to message placement were checked through component-level replay. Runtime changes were checked for state or byte equivalence, as applicable. These checks cover the tested components and operations. They do not establish identical end-to-end generated trajectories across all optimizations.

\section{Evaluation Protocol}
\label{app:eval}
\paragraph{API Cost.} Because multiple experiments ran concurrently under the same API key, we could not reliably determine the billed cost of each simulation run. We therefore estimate API costs from recorded token usage using GPT-5.6 Luna pricing at the time of the experiments, as listed in Table~\ref{tab:api_pricing}. We include all LLM calls for the reported simulation runs. Cache writes incur a 25\% surcharge over the applicable uncached input rate. Reasoning tokens are included in output usage and counted only once. Calls without usage records are disclosed separately. These estimates do not represent billed charges, and we observe that the real bills are slightly less than the estimated costs, but the difference is very marginal (about 2\% difference).

\begin{table}[t]
\centering
\small
\caption{GPT-5.6 Luna prices used for cost estimation, in USD per million tokens.}
\label{tab:api_pricing}
\begin{tabular}{lrrr}
\toprule
Input tokens per request & Input & Cached input & Output \\
\midrule
$\leq 272{,}000$ & 0.20 & 0.02 & 1.20 \\
$> 272{,}000$    & 0.40 & 0.04 & 1.80 \\
\bottomrule
\end{tabular}
\end{table}

\paragraph{Simulation Time.} We estimate generation time as the elapsed time between the earliest and latest API-call completion timestamps in each run, using the same definition across compared systems. Concurrent calls contribute only their elapsed overlap, rather than the sum of their individual durations. The interval includes request waiting, computation between calls, and logging.

\section{Scenario Initialization}
\label{app:scenario_init}

\begin{figure}[h!]
    \centering
    \includegraphics[width=\linewidth]{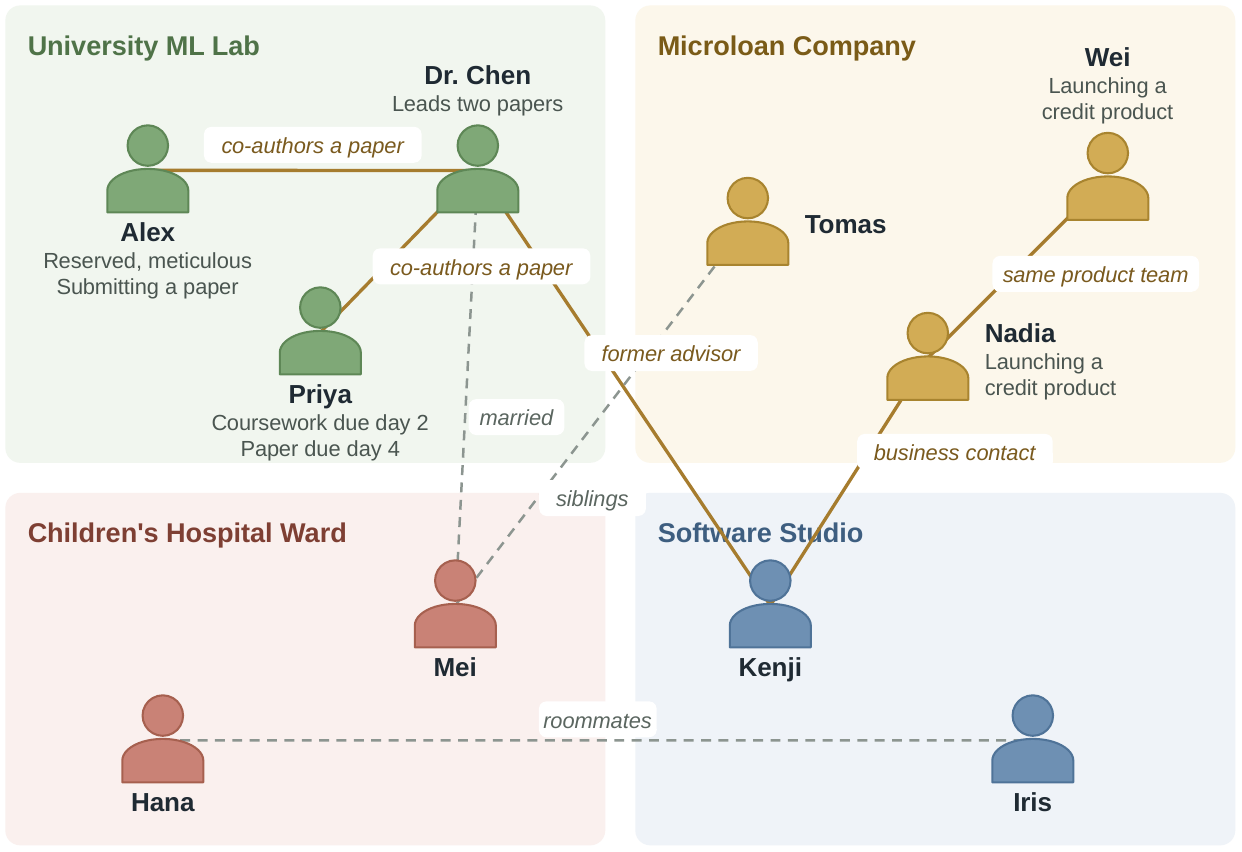}
    \caption{Initial scenario for the 60-day simulations. Ten actors occupy four connected settings, with links showing their project collaborations and other relationships. Selected actor attributes and goals are also shown.}
    \label{fig:scenario}
\end{figure}
Figure~\ref{fig:scenario} shows the initial scenario shared by all five 60-day runs. It places ten actors in four settings within one city: a university machine-learning lab (Alex, Dr. Chen, and Priya), a small lending company (Wei, Nadia, and Tomas), a children's hospital ward (Mei and Hana), and a software studio (Kenji and Iris). The actors' backgrounds connect these settings through family, housing, former advising, and work relationships. Mei is Dr. Chen's spouse and Tomas's sister, Hana and Iris share an apartment, and Kenji is Dr. Chen's former doctoral student and Nadia's work contact.

Each actor starts with a role, background, personality, stable profile, long-term life goal, and two or three initial \emph{Stage Goals}. Three shared goal groups initialize joint projects: an evidence-conflict submission for Alex and Dr. Chen, an attribution paper for Dr. Chen and Priya, and a credit-product launch for Wei and Nadia. Three seed goals have target days, including Priya's coursework on day 2 and the two attribution-paper goals on day 4. The initial world contains public facts, actor-specific visible facts, calendar entries, and setting descriptions. Before \framework{} starts, a model derives a relationship graph from the actors' backgrounds. The baselines use the same initial scenario without this derived graph.

\section{Positioning Relative to Prior Work}
\label{app:positioning}

Table~\ref{tab:positioning} provides a structured comparison of \framework{} with representative prior work along the dimensions most relevant to our setting, including long-horizon simulation, evolving actor state, persistent shared environments, multi-actor interaction, actor-specific visibility, and delayed effects.

\begin{table*}[t]
  \centering
  \caption{Positioning of E\textsuperscript{2}Sim relative to representative work.}
  \label{tab:positioning}
  \begingroup
  \fontsize{8}{9.4}\selectfont
  \setlength{\tabcolsep}{1pt}
  \renewcommand{\arraystretch}{1.12}
  \renewcommand{\tabularxcolumn}[1]{m{#1}}
  \definecolor{positioningOurs}{rgb}{0.96,0.92,0.96}
  \newcommand{\poshead}[1]{{\fontsize{8}{9}\selectfont\bfseries
    \renewcommand{\arraystretch}{1.0}%
    \begin{tabular}[c]{@{}c@{}}#1\end{tabular}}}
  \newcommand{\poswork}[2]{%
    \renewcommand{\arraystretch}{1.0}%
    \begin{tabular}[c]{@{}l@{}}#1\\[-0.4pt]
      {\fontsize{7}{8}\selectfont #2}\end{tabular}}
  \newcommand{\posyes}{\ensuremath{\checkmark}}
  \newcommand{\pospart}{\ensuremath{\sim}}
  \newcommand{\posno}{\textendash}

  \begin{tabularx}{\textwidth}{m{0.268\textwidth}|*{8}{>{\centering\arraybackslash}X}}
    \toprule
    \textbf{Work}
      & \poshead{Long\\horizon}
      & \poshead{Evolving\\actor\\state}
      & \poshead{Env.\\grounded}
      & \poshead{Persistent\\shared\\world}
      & \poshead{Multi-\\actor}
      & \poshead{Action--\\World\\feedback}
      & \poshead{Actor-\\specific\\visibility}
      & \poshead{Delayed\\effects} \\
    \midrule
    \poswork{Generative Agents}{\citep{park2023generative}}
      & \pospart & \pospart & \posyes & \posyes & \posyes & \pospart & \pospart & \posno \\[1.5pt]
    \poswork{Concordia}{\citep{vezhnevets2023generativeagentbasedmodelingactions}}
      & \pospart & \pospart & \posyes & \posyes & \posyes & \posyes & \pospart & \pospart \\[1.5pt]
    \poswork{SOTOPIA}{\citep{zhou2024sotopiainteractiveevaluationsocial}}
      & \posno & \posno & \posyes & \posno & \posyes & \pospart & \pospart & \posno \\[1.5pt]
    \poswork{LIFELONG-SOTOPIA}{\citep{goel2025lifelong}}
      & \posyes & \pospart & \posyes & \posno & \posyes & \pospart & \pospart & \posno \\[1.5pt]
    \poswork{AgentSociety}{\citep{piao2026agentsocietylargescalesimulationllmdriven}}
      & \posyes & \pospart & \posyes & \posyes & \posyes & \posyes & \pospart & \posno \\
    \midrule
    \poswork{BehaviorChain}{\citep{li-etal-2025-far}}
      & \pospart & \pospart & \pospart & \posno & \posno & \posno & \posno & \posno \\[1.5pt]
    \poswork{OmniBehavior}{\citep{chen2026realworldhumanbehaviorsimulation}}
      & \posyes & \pospart & \pospart & \posno & \posno & \posno & \posno & \posno \\
    \midrule
    \poswork{PERSONAMEM / DEEPER}{\citep{jiang2025knowmerespondme,chen-etal-2025-deeper}}
      & \posyes & \posyes & \posno & \posno & \posno & \posno & \posno & \posno \\[1.5pt]
    \poswork{LifeSim}{\citep{duan2026lifesimlonghorizonuserlife}}
      & \posyes & \posyes & \posyes & \posno & \posno & \pospart & \posno & \pospart \\
    \midrule
    \rowcolor{positioningOurs}
    \rule{0pt}{11pt}\textbf{\framework}
      & \posyes & \posyes & \posyes & \posyes & \posyes & \posyes & \posyes & \posyes \\[1pt]
    \bottomrule
  \end{tabularx}

  \vspace{4pt}
  \begin{minipage}{\textwidth}
    \fontsize{8}{9.4}\selectfont
    $\checkmark$: explicit mechanism or primary capability,\enspace
    $\sim$: partial or implicit support,\enspace
    \textendash: not a primary part of the setting.
  \end{minipage}
  \endgroup
\end{table*}

\section{Exploratory Question Answering Study}
\label{app:qa}

\begin{figure}[h!]
    \centering
    \includegraphics[width=0.56\linewidth]{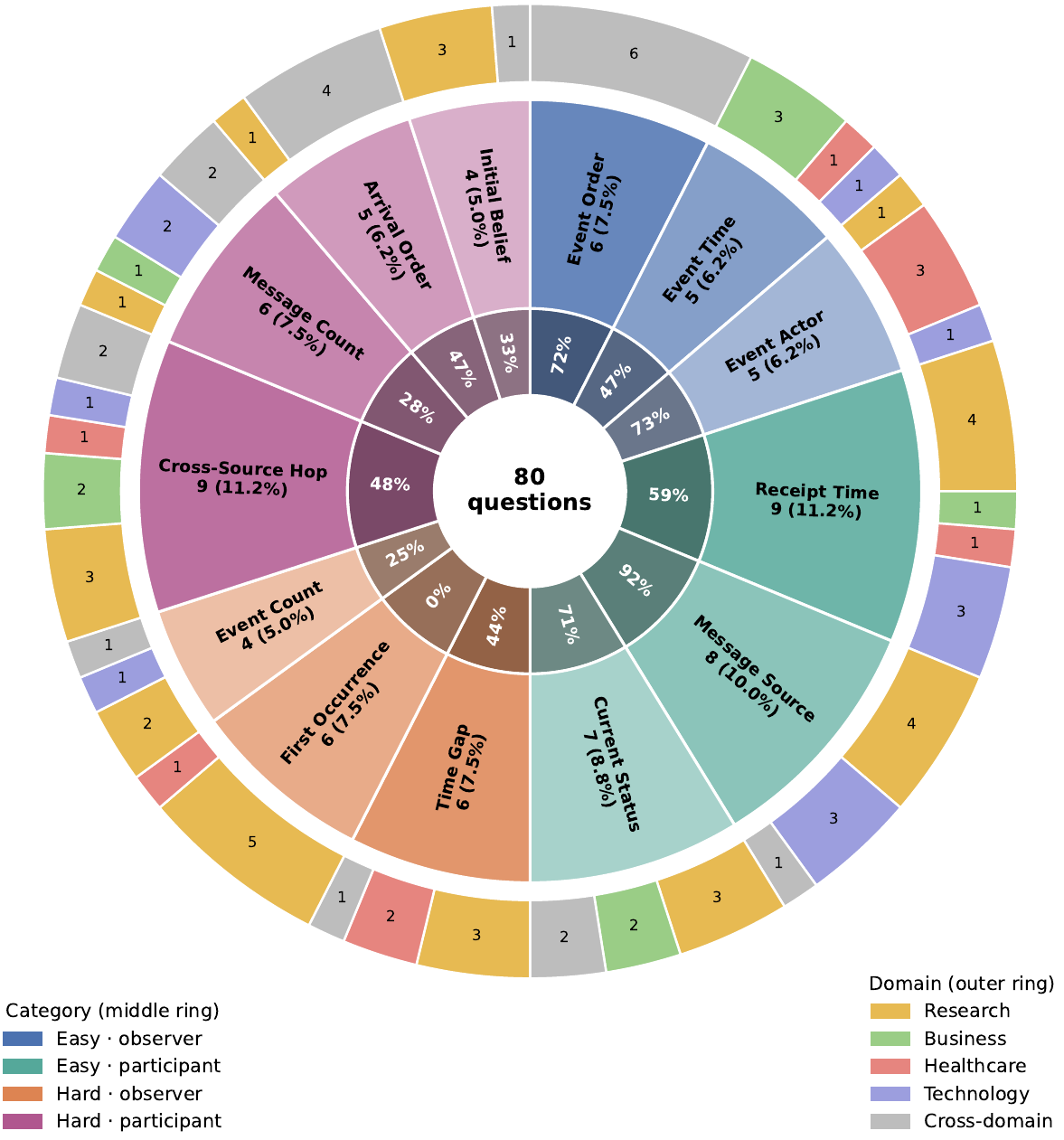}
    \caption{Question categories and scenario domains for the 80 four-choice questions. The inner ring shows \texttt{gpt-5.6-luna} accuracy with compressed memory, averaged over three answers per question.}
    \label{fig:qa_distribution}
\end{figure}

We constructed an 80-question, four-choice benchmark from the 60-day simulated trajectory to explore whether its history supports questions about actors and events. The questions cover 13 categories, with 40 easy questions answerable from a single record and 40 hard questions requiring information across longer periods or multiple actors. We checked that each answer could be recovered from the full trajectory. Figure~\ref{fig:qa_distribution} shows the question distribution.

We tested direct API question answering under two input conditions. With full history, the model receives all records up to the question time when answering each question. With compressed memory, it reads the records sequentially, updates compressed notes as new information arrives, and answers using those notes plus the most recent three days of records. This condition more closely approximates an assistant that learns about a user through ongoing interactions and must compress earlier information for later use. Each question is answered three times. Table~\ref{tab:qa_preview} reports mean accuracy. With full history, the two stronger models answer at least 90\% of hard questions, showing that most answers are recoverable. On hard questions, accuracy falls to 33--35\% when the history is processed sequentially and compressed, supporting our concern about long-horizon information use.

\begin{table}[h!]
    \centering
    \caption{Accuracy (\%) on the exploratory question-answering benchmark, averaged over three answers per question. Each difficulty level contains 40 questions.}
    \label{tab:qa_preview}
    \small
    \begin{tabular}{lcccc}
        \toprule
        & \multicolumn{2}{c}{Full history} & \multicolumn{2}{c}{Compressed memory} \\
        Model & Easy & Hard & Easy & Hard \\
        \midrule
        \texttt{gpt-6-luna} & 100.0 & 93.3 & 65.0 & 35.0 \\
        \texttt{gpt-5.6-luna} & 94.2 & 90.0 & 70.0 & 33.3 \\
        \texttt{gpt-5.4-nano} & 69.2 & 43.3 & 55.0 & 32.5 \\
        \bottomrule
    \end{tabular}
\end{table}

This is a preliminary test on questions from one synthetic trajectory. The answers are produced by direct API calls to language models, without an acting agent or an agentic retrieval workflow. The scores therefore measure question-answering accuracy under the specified inputs, not the performance of a deployed assistant agent.

\end{document}